\documentclass{article} % For LaTeX2e
\usepackage{iclr2025_conference,times}

\usepackage{amsmath,amsfonts,bm}

\def\eqref#1{equation~\ref{#1}}
\def\Eqref#1{Equation~\ref{#1}}
\def\1{\bm{1}}

\DeclareMathAlphabet{\mathsfit}{\encodingdefault}{\sfdefault}{m}{sl}
\SetMathAlphabet{\mathsfit}{bold}{\encodingdefault}{\sfdefault}{bx}{n}

\usepackage[utf8]{inputenc} % allow utf-8 input
\usepackage[T1]{fontenc}    % use 8-bit T1 fonts
\usepackage{url}            % simple URL typesetting
\usepackage{booktabs}       % professional-quality tables
\usepackage{amsfonts}       % blackboard math symbols
\usepackage{nicefrac}       % compact symbols for 1/2, etc.
\usepackage{microtype}      % microtypography
\usepackage{xcolor}         % colors
\usepackage{float}
\usepackage{natbib}
\usepackage[ruled,linesnumbered,noend]{algorithm2e} % linesnumbered

\newcommand{\eg}{{e.g.}\xspace}
\newcommand{\Dcal}{\mathcal{D}}
\newcommand{\ie}{{i.e.}\xspace}
\newcommand{\Ucal}{\mathcal{U}}
\newcommand{\xvec}{\vec{x}}
\newcommand{\shat}{\hat{s}}

\usepackage{microtype}
\usepackage{graphicx}
\usepackage{multicol}
\usepackage{booktabs} % for professional tables
\usepackage{multirow}
\usepackage[inline,shortlabels]{enumitem}

\usepackage{amsmath}
\usepackage{amssymb}
\usepackage{mathtools}
\usepackage{amsthm}
\usepackage{subcaption}
\usepackage{caption}

\usepackage[colorlinks,linkcolor=red,citecolor=blue,urlcolor=black]{hyperref}
\usepackage[capitalize,noabbrev]{cleveref}
\usepackage[textsize=tiny]{todonotes}

  {\list{}{\leftmargin=0.3in\rightmargin=0.3in}\item[]}%
  {\endlist}
\makeatletter
\patchcmd{\@algocf@start}% <cmd>
  {-1.5em}% <search>
  {0pt}% <replace>
  {}{}% <success><failure>
\makeatother

\title{Augmenting Offline Reinforcement Learning with State-only Interactions}

\author{Shangzhe Li\thanks{This work was done during an internship in Department of Computer Science, University of Illinois Chicago.}~, Xinhua Zhang \\
University of Illinois Chicago\\
\texttt{\{shanli3,zhangx\}@uic.edu} \\
}

\iclrfinalcopy % Uncomment for camera-ready version, but NOT for submission.
\begin{document}

\maketitle

\begin{abstract}

Batch offline data have been shown considerably beneficial for reinforcement learning.
Their benefit is further amplified by upsampling with generative models.
In this paper, we consider a novel opportunity where interaction with environment is feasible, 
but only restricted to observations,
\ie, \textit{no reward} feedback is available.
This setting is broadly applicable,
as simulators or even real cyber-physical systems are often accessible,
while in contrast reward is often difficult or expensive to obtain.
As a result, the learner must make good sense of the offline data to synthesize an efficient scheme of querying the transition of state. 
Our method first leverages online interactions to generate high-return trajectories via conditional diffusion models.
They are then blended with the original offline trajectories through a  stitching algorithm,
and the resulting augmented data can be applied generically to downstream reinforcement learners.
Superior empirical performance is demonstrated over state-of-the-art data augmentation methods that are extended to utilize state-only interactions.
\end{abstract}

% \vspace{-0.5em}
\section{Introduction}
% \vspace{-0.1em}

Data augmentation has long been an effective approach that boosts the performance of learning algorithms.
% For example, mix-up interpolates training examples to promote linear behavior between examples \citep{}.
They have been particularly useful for neural networks whose regularization and generalization properties are much harder to characterize than shallow models.
In reinforcement learning (RL),
experience replay \citep{fedus2020revisiting,mnih2015human} can be considered as augmenting the latest experiences with the past ones when updating the model.
%
% Online interaction with the environment underpins conventional reinforcement learning (RL) methods.
% However, they can be time-consuming and expensive.
% This issue was alleviated by experience replay [19, 53], 
% which augments the latest experiences with the past ones when updating the model.
%
An even more effective paradigm is offline RL \citep{levine2020offline},
where a batch of previously collected trajectories are used to boost online RL.
They can be used to augment the online update buffer, 
or to pre-train a model that is subsequently fine-tuned by online RL.

% Offline RL \citep{levine2020offline}, which is typically fine-tuned online,
% can be considered an even more aggressive augmentation,
% using a batch of previously collected trajectories.
% Sometimes, they perform well even without online fine-tuning.

\begin{figure*}[t]
%\vskip 0.2in
\begin{center}
\centerline{\includegraphics[width=\textwidth]{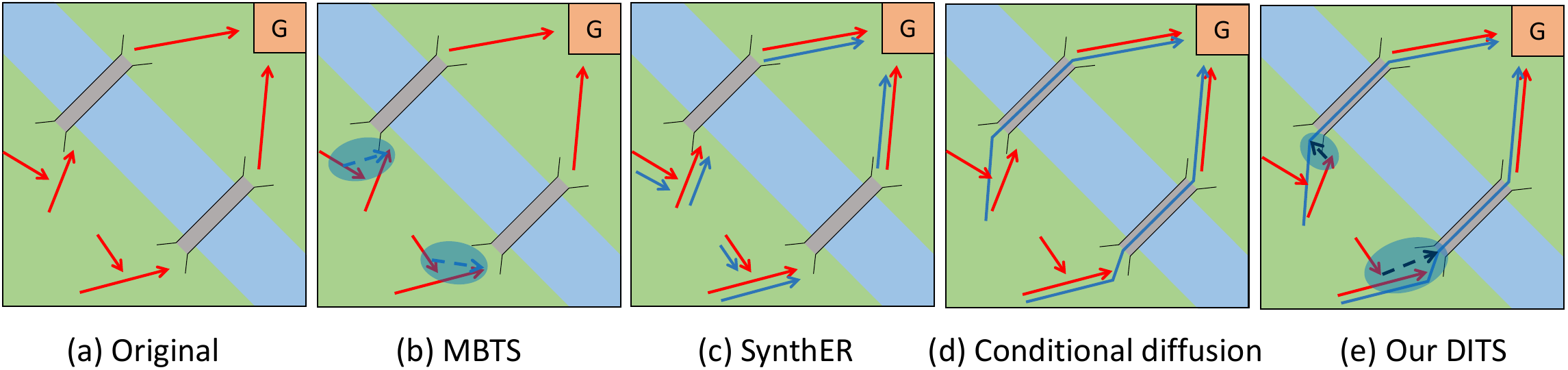}}
\vskip -0.2em
\caption{An example that illustrates the difference between four data augmentation methods. 
\textbf{(a)}: the original offline dataset, 
where a river is in the main diagonal with two bridges.  
There are six random trajectories.  
Two of them walk to the Goal state (top right), 
which gives a high reward.
Four trajectories are stuck at the lower-left half with a low reward.
\textbf{(b)}: MBTS concatenates trajectories, 
but they are restricted to original trajectories,
instead of leveraging the bridge-crossing one that could be synthesized by a generative model.
\textbf{(c)}: the SynthER method where all transitions make an equal contribution to the training of the diffusion model.
\textbf{(d)}: the conditional diffusion model which super-samples high-return trajectory regions.
As a result, new trajectories that cross the bridges are formed.
\textbf{(e)}: our DITS method of conditional diffusion followed by stitching, 
which concatenates low-return trajectories with high-return ones.
In consequence, two dashed arrows are formed allowing two low-return trajectories to be connected to a bridge-crossing one.
}
\label{fig:illustration}
\end{center}
\vskip -2.7em
\end{figure*}

Offline RL may suffer from sub-optimal batch data because the behavior policy could be sub-optimal,
and the data may not have sufficiently covered the environment,
especially the high-rewarding regions.
To address this issue,
stitching approaches have been studied \citep{li2024DiffStitch}.
Model-based trajectory stitching \citep[MBTS,][]{Hepburn24} augments the offline dataset by connecting low-reward trajectories with high-reward ones.
% They claimed to stitch high-valued regions together in the paper. Therefore, in our case, it's stitching a high-valued state from a section with overall poor demonstrations to the high-return trajectories.
% If the section of demonstration has no high-valued state, then this part will be discarded.
This is shown in Figure~\ref{fig:illustration}b,
where two dashed lines are added,
connecting to a point that is closer to a bridge (higher value).
However, both \citet{li2024DiffStitch} and \citet{Hepburn24} only stitch \textit{existing} trajectories
while no novel high-reward ones are generated.

This issue appears solvable by generative models.
As shown in Figure~\ref{fig:illustration}c, 
SynthER augments the set of \textit{transitions} by training a diffusion model \citep{lu2023synthetic}.
However, it does not generate \textit{trajectories} which would require auto-regressive models.
In our experiment, directly doing so performs worse than SynthER which upsamples transitions,
suggesting that as a data augmentation approach to RL,
auto-regressively generating trajectories may accumulate high bias and produce low return.

To tackle this problem,
we make a new observation that some limited form of interactions are helpful.
Indeed,  online \textit{state-only} interaction with environment is feasible in many applications.
For example:
\begin{itemize}[leftmargin=*]
\setlength\itemsep{0em}
    \item RL has been adopted to improve treatment policy of chronic disease by factoring in the delayed effects of treatment \citep{weltz2022reinforcement}.
Many digital twins are available that simulate a patient's condition \citep{tardini2022optimal}.
However, toxicity feedback (part of the reward)
may exhibit marked variance across patients,
both physically and mentally.
So it requires measurement on real persons from offline recordings.
    \item In inventory control, multi-product stock levels can often be well simulated for a stocking policy \citep{chen2024learning}.
However, the real reward (profit) also depends on the price which in turn fluctuates with the market and the marketing strategy (\eg, discount).
So it is generally difficult to retrieve rewards in an online setting, 
and we need to best utilize the reward from the offline data.
    \item In sequential recommendation, it is common to model a user's behavior such as click-through rate. 
    But the alignment with their ethical principle (reward) cannot be automatically inferred online \citep{stray2021what}.
\end{itemize}

\vspace{-0.2em}
%
% However, it is often difficult to design reward functions that effectively facilitate RL.
% As a result, imitation learning has been well studied which learns from demonstrator's behavior in the absence of reward feedback.
The question we investigate and answer in the affirmative is:
%
% \begin{quote}
    \textbf{can we improve trajectory generation, as a data augmentation approach for RL, by making \textit{efficient} use of state-only interactions?}%
    \footnote{We work on state interactions here,
leaving the extension of observation-only interactions to future work.}
% \end{quote}
%
% The efficiency requirement demands judicious selection of state-action pairs to interact with the environment,
% and ideally, the movement should be smooth,
% such that the simulator or real robots do not have to be reset to a very distant state at every step.
% We would naturally prefer to reserve such interaction opportunities to high-reward regions.
%
A natural approach is to learn a good policy from offline RL,
and deploy it by interacting with the environment to obtain a sequence of state and action.
Finally, impute the reward from state and action via a regression model trained from offline data.
Within this framework, our major contribution lies in two folds.
Firstly, we make the key observation that although synthesizing high-return trajectories ameliorates the possible deficiency of such trajectories in the offline data,
it may also lead to overfitting of such regions.
Therefore, we blend them into the original dataset progressively with a \textit{selective} stitching strategy,
introducing new transitions between two states on \textit{different} trajectories.

Secondly, we identify a policy model that best fits our setting.
Since the interaction is not at test time,
the policy's efficiency is not a main concern,
allowing us to leverage Transformer or diffusion based planners.
In particular, we extended the decision diffuser \citep[DD,][]{Ajay23},
which trains a diffusion model to generate trajectories \textit{conditioned} on high return.
For example, in Figure~\ref{fig:illustration}d,
new trajectories are generated that cross the bridge to reach the goal state.
Moreover, DD diffuses state sequences,
which is exactly what is collected from new interactions.
Such an alignment facilitates additional online fine-tuning of the diffuser;
we leave it for future work.

We will refer to our method as Diffusion-based Trajectory Stitching  (DITS).
As shown in Figure~\ref{fig:illustration}e,
DITS enjoys the benefit of conditional diffuser in generating new high-rewarding trajectories,
and it also fixes the limitation of MBTS by allowing low-reward trajectories to connect with the high-reward ones generated by the diffuser.
From the trajectory generation perspective,
DITS circumvents auto-regression,
delegating the backbone progression of state to direct interaction,
reducing the task of trajectory generation to step-wise synthesis of actions and rewards.
We will also show that both of them can be accomplished by DITS via a single diffusion model.

\begin{figure*}[t]
%\vskip 0.2in
\begin{center}
\centerline{\includegraphics[width=\textwidth]{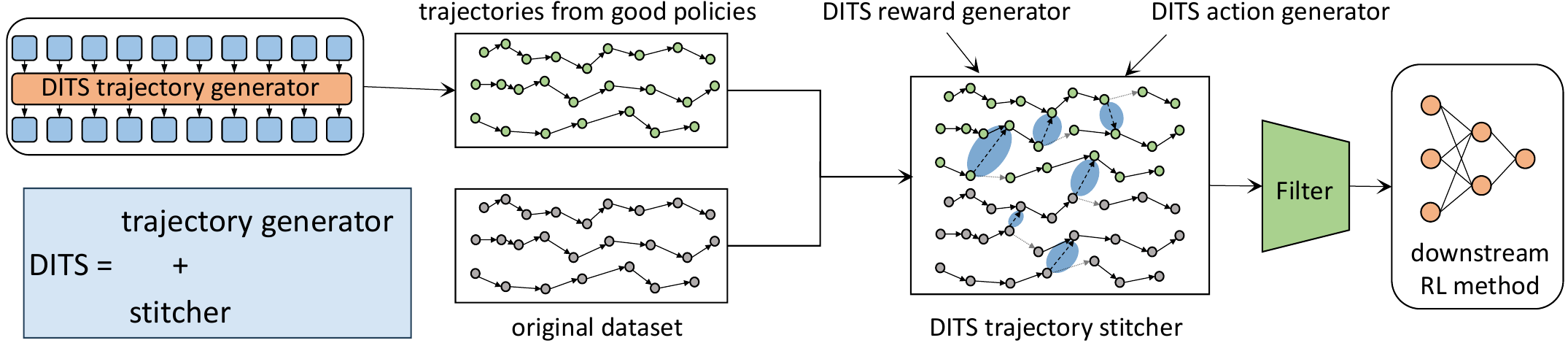}}
\vskip -0.2em
\caption{\textbf{Di}ffusion-based \textbf{T}rajectory \textbf{S}titching (\textbf{DITS}). Trajectories from DITS' trajectory generator are combined with the original dataset for the stitching process, which creates new transitions (blue arrows in the ``Trajectory stitching'' block) and discards old transitions (grey dotted edges). 
Stitching was facilitated by DITS' reward and action generators. 
A filter is applied to prune away low-return trajectories, 
and the result is finally used by a downstream RL method.}
\label{fig:TSKD}
\end{center}
\vskip -2.2em
\end{figure*}

The workflow of DITS is also charted in Figure~\ref{fig:TSKD}.
The offline dataset is first used to train DITS' generator,
producing trajectories that are blended with the original dataset through DITS' stitcher.
The stitcher progressively evaluates the candidate states to transition to by using a series of criteria,
leveraging the reward and actions generated by DITS.
The resulting pool is filtered based on the full trajectory return,
and the remaining trajectories are supplied to a downstream RL algorithm.

This paper is organized as follows.
After the preliminaries in Section \ref{sec:prelim},
we introduce the conditional decision diffuser used by DITS' generator, 
which produces full trajectories conditioned on high returns (Section~\ref{sec:ddr}).
% and DDR-II which generates the reward given the current and next states.
Based on it,
% these constructs, 
the stitching algorithm is presented in Section~\ref{sec:TSKD},
and its superior performance over other data augmentation methods is empirically demonstrated in Section~\ref{sec:exp}.

% We \textbf{emphasize} that our method aims to leverage state-only interactions from a data augmentation perspective,
% proffering generic applicability to RL algorithms.
% There are obviously other ways of utilizing interactions,
% which can be further customized for an RL method.
% We leave it to future work.

% \citep{schmeckpeper2021reinforcement}

\iffalse
Offline reinforcement learning (RL) presents a novel opportunity to leverage offline batch data without having to interact with the environment \citep{levine2020offline}. 
Simple approaches such as Behavioral Cloning (BC) are directly applicable,
minimizing the error between the target policy and the learned policy. 
However, both BC and many well-established \textit{online} RL methods suffer from the out-of-distribution (OOD) issues because once deployed online, 
the state distribution can shift considerably from that in the offline data,
and the Q-value can be over-estimated on OOD states-actions
\citep{fujimoto2019offpolicy,kumar2019stabilizing,fu2019disgnosing,kumar20discor}.

A large number of methods have been proposed to address this issue,
and we only highlight a few.
%
\citet{kumar2020conservative} proposed conservative Q-learning (CQL) to pessimistically under-estimate the state-action values for OOD actions.
\citet{nair2020awac} and several other works constrain the policy to the proximity of the behavior policy.
Implicit Q-learning \citep[IQL,][]{Kostrikov21} avoids querying values of unseen actions while performing multi-step dynamic programming updates. 

\fi

\vspace{-0.4em}
\paragraph{Related Work}

Diffusion based models have been recently shown effective for  RL.
Diffusion policy models \citep{chi2023diffusionpolicy,wang2023diffusion,ze2024diffusion} refine noise into actions through a gradient field.
It is conditioned on the state or observation,
and can be stably trained.
The learned policy can accurately model multimodal and high-dimensional action distributions.
In contrast, DD diffuses on state spaces instead of actions,
and generates the action via an inverse dynamics model.
It also allow more general conditioning such as high reward, constraint satisfaction, and skill composition.
\citet{Wang23} proposed augmenting behavioral cloning by diffusion over state and action pairs,
but it does not consider conditioning.

\vspace{-0.3em}
\section{Preliminary}
\label{sec:prelim}

We follow the standard setting of Markov decision process \citep[MDP,][]{Puterman95},
which is defined by the tuple 
$\langle \rho_0, \mathcal{S}, \mathcal{A}, \mathcal{T}, \mathcal{R}, \gamma\rangle$.
Here $\rho_0$ is the initial distribution. 
$\mathcal{S}$ is the $m$-dimensional state space,
and $\mathcal{A}$ is the action space.
$\mathcal{T}: \mathcal{S}\times\mathcal{A}\rightarrow \Delta_\mathcal{S}$ is the transition function.
$\mathcal{R}: \mathcal{S}\times\mathcal{A}\times\mathcal{S}\rightarrow\mathbb{R}$ is the reward function.
$\gamma\in [0, 1)$ is a discount factor. 
The agent in this environment acts with a policy $\pi:\mathcal{S}\rightarrow \mathcal{A}$, 
generating a sequence of state-action-reward transitions, 
which represents a trajectory $\tau\coloneqq(s_t, a_t, r_t)_{t\geq0}$ with probability $p_\pi(\tau)$. 
The return of the trajectory is $R(\tau) \coloneqq \sum_{t\geq0}\gamma^tr_t$.

An offline dataset $\Dcal$ consists of trajectories $(\tau_1, \ldots, \tau_T)$ which are collected using a behavior policy $\pi_\beta$.
Each trajectory is composed of a set of transitions $\{(s_t, a_t, r_t, s'_t)\}$,
where we customarily use $'$ to denote the \textit{next} state,
\ie, $s'_t = s_{t+1}$.
SynthER \citep{lu2023synthetic} directly builds a generative model based on the union of such transition sets across all the $T$ trajectories to replay the experience.

\iffalse
\vspace{-0.4em}
\paragraph{Behavioural Cloning}

% We are interested in the algorithm of Behavioural Cloning (BC), which performs imitation learning in offline RL setting. 
A BC agent learns a policy $\pi(a|s)$ 
% that maximizes the returns of the trajectories by 
by performing supervised learning on the dataset sampled from the environment $\mathcal{D} = \{(s_t, a_t, r_t, s'_t)\}$.
Here, $s'_t$ denotes the next state for the transition. 
BC seeks the optimal policy $\pi_\theta$ parameterized with $\theta$ by minimizing the risk
\begin{equation*}
    \mathcal{L_{BC}} \coloneqq \mathbb{E}_{s, a\sim\mathcal{D}} [(\pi_\theta(s)-a)^2].
\end{equation*}
%
Extension to stochastic policy is straightforward,
and we just keep the exposition simple because our work focuses on  augmenting the offline dataset that is used for BC training.

BC can be implemented easily by parameterizing the policy with a Multilayer Perceptron (MLP), 
using a relatively small amount of parameters. 
However, BC performs poorly when the quality of the transitions in the dataset is not high. 
Therefore, in this work we aim to perform knowledge distilling from a deep generative model of the trajectories, 
which in our case is a decision diffuser with reward, 
to extract high-return trajectories to augment the dataset through a nontrivial stitching process. 
As a result, the policy learned from BC can be both effective and simple,
representing an efficacious distillation from the deep generative model.
\fi

\vspace{-0.6em}
\paragraph{Diffusion Probabilistic Models (DDPM)}

Diffusion models \citep{Ho20} are a class of generative models that learn the data distribution $q(\mathbf{x})$ from a dataset $\mathcal{D}\coloneqq\{\mathbf{x}^i\}_{0\leq i<M}$. One of the major usage of these models is generating high-quality images from text descriptions \citep{Saharia22}. The diffusion models consist of two processes, which are the forward process and the reverse process. For the forward process, the model will add noise step by step via a predefined way $q(\mathbf{x}_{k+1}|\mathbf{x}_{k})\coloneqq\mathcal{N}(\mathbf{x}_{k+1}; \sqrt{\alpha_k}\mathbf{x}, (1-\alpha_k)\mathbf{I})$. For the reverse process, the model has trainable parameters in order to learn a good way to denoise step by step, which can be represented mathematically as $p_\theta(\mathbf{x}_{k-1}|\mathbf{x}_k)\coloneqq\mathcal{N}(\mathbf{x}_{k-1}|\mu_\theta(\mathbf{x}_k, k), \Sigma_k)$. In these processes, $\mathcal{N}(\mu, \Sigma)$ denotes a Gaussian distribution with mean $\mu$ and covariance matrix $\Sigma$, and $\alpha_k\in\mathbb{R}$ determines the variance schedule. $\mathbf{x}_0\coloneqq\mathbf{x}$ denotes the original sample, while $\mathbf{x}_1$, $\mathbf{x}_2$, ..., $\mathbf{x}_{K-1}$ are the intermediate steps for the diffusion, and $\mathbf{x}_K\sim\mathcal{N}(\mathbf{0}, \mathbf{I})$ is the unit Gaussian noise.

The reverse process model can be trained by minimizing the following loss \citep{Ho20}:
\begin{equation*}
    \mathcal{L}_{denoise}(\theta)\coloneqq\mathbb{E}_{k\sim \Ucal[1, K], \mathbf{x}_0\sim q,\epsilon\sim\mathcal{N}(\mathbf{0}, \mathbf{I})}[\Vert\epsilon-\epsilon_\theta(\mathbf{x}_k, k)\Vert^2].
\end{equation*}

The noise model $\epsilon_\theta(\mathbf{x}_k, k)$ is parameterized by a deep temporal U-Net \citep{Janner22}. 
This is equivalent to modeling the mean of $p_\theta(\mathbf{x}_{k-1}|\mathbf{x}_k)$,
as $\mu_\theta(x_{k-1}|x_k)$ can be calculated from $\epsilon_\theta(\mathbf{x}_k, k)$ \citep{Ho20}.

% \vspace{-1em}
\paragraph{Guided Diffusion}
The diffusion training process can be guided using conditional data. In order to avoid the need of a strong classifier, classifier-free guidance method \citep{Ho22} is preferred. 
This method learns both a conditional model $\epsilon_\theta(\mathbf{x}_k, \mathbf{y}, k)$ and an unconditional model $\epsilon_\theta(\mathbf{x}_k, \emptyset, k)$, 
where $\mathbf{y}$ is the label and $\emptyset$ is a dummy value that takes the place of $\mathbf{y}$ for unconditional situations. 
The perturbed noise can be denoted as 
$\epsilon_\theta(\mathbf{x}_k, \emptyset, k) + \omega(\epsilon_\theta(\mathbf{x}_k, \mathbf{y}, k) - \epsilon_\theta(\mathbf{x}_k, \emptyset, k))$, 
where $\omega$ denotes the guidance scale or the trade-off factor.

\iffalse
\vspace{-0.4em}
\paragraph{Knowledge Distillation}
Knowledge distillation (KD) transfers knowledge from one deep learning model (the teacher) to another (the student). 
It was originally proposed by minimizing the KL divergence of the outputs between the teacher model and the student model \citep{Hinton15}.

Our work proposes a distillation framework that transfers the knowledge of a well-learned policy based on deep generative models to a relatively shallow model such as BC network by augmenting the dataset via trajectory stitching.
\fi

\vspace{-0.23em}
\section{Trajectory Generator of DITS with State-only Interaction}
\label{sec:ddr}
\vspace{-0.2em}

We first present our trajectory generator,
which produces $(s_t, a_t, r_t, s'_t)$ sequences through state-only interactions.
A vanilla Decision Diffuser \citep[DD,][]{Ajay23} works as a planner that proposes actions \textit{conditioned} on high reward or other constraints,
based on which the state-only interaction produces the next state $s'_t$.
However, DD does not generate rewards,
while many applications lack a reward formula,
\eg, patient's mental stress level.
We thus improve DD by also generating rewards.

% and an inverse dynamics model (IDM). 
% In real applications, most environments do not employ a predefined reward function. 
% Therefore, given any offline RL algorithm that learns from an offline dataset with rewards $\Dcal$,
% any data augmentation approach should be able to generate full trajectories that includes rewards.
% Therefore, compared with the original Decision Diffuser, 
% our model is able to generate not only states and actions but also rewards. 
% Compared with knowledge distillation methods of goal-conditioned $Q$-learning \citep{Levine23}, 
% our DDR model does not require estimating any kind of $Q$-function. 
% Moreover, 
% our model does not suffer the risk of distribution-shift because the generative models are trained with maximum-likelihood estimations \citep{Ajay23}. 

Analogously to \citet{Ajay23},
we only diffuse on the states and rewards,
because the sequence over actions, 
which are often represented as joint torques, 
tend to have higher frequency and are less smooth, 
making them much harder to model \citep{Tedrake22}. 
We therefore decide to model the sequence of $(s_t, r_t, s'_t)$ by a diffusion model,
and train an inverse dynamics model (IDM) to infer $a_t$ from $s_t$ and $s'_t$ (Section~\ref{sec:IDM}).

\vspace{-0.2em}
\subsection{Conditional Diffusion Models for Decision Making}

We can formulate sequential decision-making as a problem of conditional generative modeling:
\begin{align}
\label{eq:obj_cond_diffuser}
    \max\nolimits_\theta \ 
    \mathbb{E}_{\tau\sim\mathcal{D}}
    [\log p_\theta(\mathbf{x}_0(\tau)|\mathbf{y}(\tau))].
\end{align}
Our aim is to generate a \textit{partial} trajectory $\mathbf{x}_0(\tau)$ using the information of conditioning data $\mathbf{y}(\tau)$. 
$\mathbf{x}_0(\tau)$ can be any subsequence of $\tau$,
and $\mathbf{y}(\tau)$ can be the corresponding return, 
the constraints satisfied by it, 
or the skill demonstrated in it. 
Since DITS aims to generate high return trajectories,
we will use the return $\sum_{i=t}^{t+H}\gamma^{i-t} r_i$ as our condition $\mathbf{y}(\tau)$,
when $\xvec_0(\tau)$ assumes a subsequence of $\tau$ from $t$ to $t+H$.
Following \citet{Ajay23},
we omit writing out the expectation over the random subsequence of $\tau$ in \Eqref{eq:obj_cond_diffuser}.

We can then construct our generative model according to the conditional diffusion process:
\begin{equation}
    q(\mathbf{x}_{k+1}(\tau)|\mathbf{x}_{k}(\tau)),
    \quad 
    p_\theta(\mathbf{x}_{k-1}(\tau)|\mathbf{x}_{k}(\tau), \mathbf{y}(\tau)).
\end{equation}
Here, $q$ represents the forward diffusion process while $p_\theta$ represents the trainable reverse process parameterized by $\theta$. 
Compared with the original Diffusion Probabilistic Model \citep{Ho20}, 
the reverse process here has an additional conditioning label $\mathbf{y}(\tau)$, 
% which in our case is the return of the trajectory, 
guiding the reverse process with high return condition and enabling the model to learn a good policy in the environment.

\begin{figure*}[t]
% \vskip 0.2in
\begin{center}
%\vskip -0.5in
\centerline{\includegraphics[width=\textwidth]{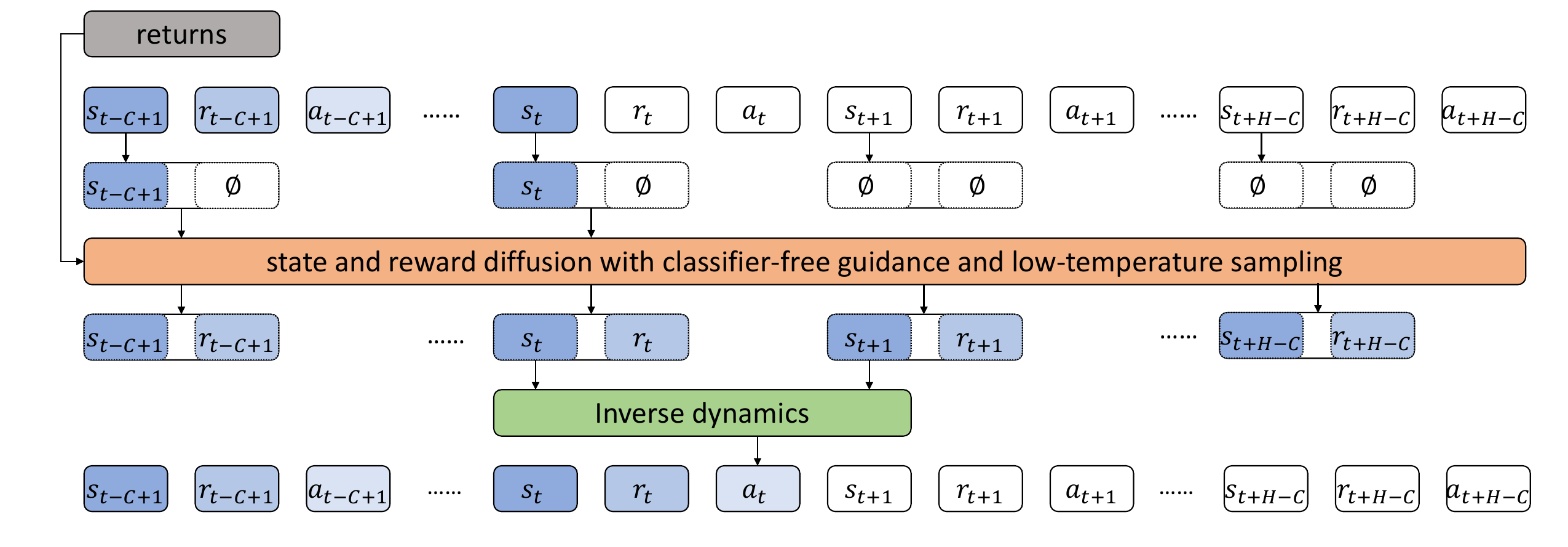}}
\vskip -0.2em
\caption{\textbf{Trajectory generation of DITS}. Given the latest $C$ number of states $s_{t-C+1},\ldots, s_t$, the diffuser uses classifier-free guidance with low-temperature sampling to generate a sequence of future states and rewards. 
IDM is applied to generate action $a_t$ with $s_t$ and $s_{t+1}$.
$\emptyset$ means the value is left for the diffuser to fill in, instead of clamping with the observation (if available).}
\label{DDR-I}
\end{center}
\vskip -2em
\end{figure*}

\vspace{-0.2em}
\subsection{Trajectory generation in DITS via state-only interaction}

\begin{algorithm}[t]
   \caption{Trajectory generation in DITS via state-only interaction with a 
\textit{receding-horizon} control}
   \label{alg:DDR-I}
   \textbf{Input}: Noise model $\epsilon_\theta$, inverse dynamics $f_\phi$, guidance scale $\omega$, history length $C$, condition $\mathbf{y}$.
   
   {\color{red} Initialize $h\leftarrow Queue(length = C)$}. 
   Initialize state $s_0$ and an empty trajectory $\tau$. 
      
   \For{$t = 0, 1, ...$ until done}{
   
   % \underline{\textbf{Observe state} $s$} and 
   {\color{red} $h.insert(s_t)$}.
   % Set $s_t \! \gets \! s$. 
   Initialize $\mathbf{x}_K(\tau) \! \sim \! \mathcal{N}(\mathbf{0}, \alpha\mathbf{I})$ in $\mathbb{R}^{H(m+1)}$ by \Eqref{eq:xktau_def_ddr_1}. 
   
   \For{$k = K...1$} {   
   {\color{red} $\mathbf{x}_k(\tau)[:length(h)].\text{states} \gets h$}.
   % \IF{$k=K$}
   % \STATE $h_{cond} \leftarrow h$
   % \ELSE
   % \STATE Extract $r_k$ from $x_k(\tau)$
   % \STATE Replace all $r$ in $h_{cond}$ with $r_k[:length(h_{cond})]$
   % \ENDIF
   % \STATE 
   % $\mathbf{x}_k(\tau)\texttt{[}:length(h_{cond})\texttt{]}\leftarrow h_{cond}$ 
   %
   Set $\hat \epsilon$ with Equation \ref{classifier-free guidance}, using $\theta$, $\mathbf{x}_k(\tau)$, $\mathbf{y}$, and $k$. 
   % $\hat{\epsilon}\leftarrow\epsilon_\theta(\mathbf{x}_k(\tau), \emptyset, k) + \omega(\epsilon_\theta(\mathbf{x}_k(\tau), \mathbf{y}, k) - \epsilon_\theta(\mathbf{x}_k(\tau), \emptyset, k))$
        
   Sample $\mathbf{x}_{k-1}(\tau)\sim\mathcal{N}(\mu_{k-1}, \alpha\Sigma_{k-1})$,
   where $(\mu_{k-1}, \Sigma_{k-1})\leftarrow \text{Denoise}(\mathbf{x}_k(\tau), \hat\epsilon)$.
   
   }
   $r_t \! \gets \! \mathbf{x}_0(\tau)[C\!-\!1].\text{reward}$.    
   \hspace{0.2em} 
   $s_{t+1} \! \gets \! \mathbf{x}_0(\tau)[C].\text{state}$. 
   \hspace{0.2em} 
   $a_t \gets f_\phi(s_t, s_{t + 1})$ from IDM.

   {\color{blue} Apply $a_t$ to the environment}.
   Overwrite $s_{t+1}$ with the true next state.
   % \hspace{0.5em} and \hspace{0.5em} 
   $\tau.\text{append}(s_t, a_t, r_t, s_{t + 1})$.
   }
\end{algorithm}

We will diffuse over the following state for diffusion timesteps $k = 1, \ldots, K$:
\vspace{-0.2em}
\begin{align}
\label{eq:xktau_def_ddr_1}
    \mathbf{x}_k(\tau) &\coloneqq (s_{t-C+1}, r_{t-C+1}, \ldots, s_t, r_t, \ldots, s_{t+H-C}, r_{t+H-C})_k 
    \ \in \mathbb{R}^{{\color{blue}H(m+1)}}.
\end{align}
\vspace{-1.4em}

%Here, $t$ denotes the step index in a trajectory.
In the process of diffusion,
it is crucial to make the current part of the trajectory consistent with the history. 
We therefore introduce the most recent $C$ number of states by repeatedly overwriting the entries of $s_{t-C+1}, \ldots, s_{t}$ in $\mathbf{x}_k(\tau)$,
throughout all diffusion steps $k$.
We finally extract $r_t$ and $s_{t+1}$ from $\mathbf{x}_0(\tau)$.
If $s_{t+1}$ is used directly to run the denoising process again, 
we end up with an auto-regressive generation of trajectories,
which turned out ineffective in our experiments.

DD trains an IDM to infer the action $a_t$ from $s_t$ and $s_{t+1}$,
which can be used to interact with the environment and to obtain the true next state $s_{t+1}$.
The whole generation procedure is formalized in Algorithm~\ref{alg:DDR-I}
and is illustrated in Figure~\ref{DDR-I}.
We will refer to the entry of $s_{t-C+1}$ as $\mathbf{x}_{k}(\tau)[0].\text{state}$,
and the entry of $r_{t-C+1}$ as $\mathbf{x}_{k}(\tau)[0].\text{reward}$.

\phantom{a}

\vspace{-2.3em}
It is noteworthy that we do not write in the reward from the latest $C$ steps, 
differing from the states.
Instead, we just let the diffuser fill in their values,
which is consistent with our postulation that past states provide sufficient information to predict the reward.
Further, although we do not pursue it in this work,
our generator could also be used in online model-based planning,
in which case the past rewards may not be even available to the agent online.

\vspace{-0.2em}
\subsection{Conditioning with Classifier-free Guidance and the Training Objective}

To train the generator,
we employ classifier-free guidance to integrate conditional influence into the diffusion process \citep{Ho22}. 
This is superior to using a classifier that requires estimating the $Q$-function \citep{Ajay23}. 
It defines a perturbed noise that is applied at trajectory generation:
\vspace{-1em}
\begin{align}
\label{classifier-free guidance}
\hat{\epsilon}\coloneqq\epsilon_\theta(\mathbf{x}_k(\tau), \emptyset, k) + 
\omega \cdot [&\epsilon_\theta(\mathbf{x}_k(\tau), \mathbf{y}(\tau), k) - \epsilon_\theta(\mathbf{x}_k(\tau), \emptyset, k)],
\end{align}
where the scalar $\omega$ tends to augment and extract the best portions of trajectories in the dataset that comply with the conditioning of $\mathbf{y}(\tau)$, i.e., high return here.
This helps the decision diffuser to learn a good policy from an average dataset. 
Then the reverse process $p_\theta$, 
which is parameterized by the noise model $\epsilon_\theta$,
can be learned by minimizing \citep{Ho22}
\vspace{-0.1em}
\begin{align}
    &\mathcal{L}_{\text{gen}}(\theta)\coloneqq \mathbb{E}_{\epsilon \sim\mathcal{N}(\mathbf{0}, \mathbf{I}), 
    k \sim\mathcal{U}\{1, ..., K\}, \tau\in\mathcal{D}, \beta\sim \text{Bern}(p)} [\Vert\epsilon-\epsilon_\theta(\mathbf{x}_k(\tau), (1 - \beta)\mathbf{y}(\tau)+\beta\emptyset, k)\Vert^2 ].
\end{align}

% For each trajectory $\tau$, we first sample a noise $\epsilon\sim\mathcal{N}(\mathbf{0}, \mathbf{I})$ and a timestep $k\sim\mathcal{U}\{1, ..., K\}$. 
% Then, we construct the noisy array of $\mathbf{x}_k(\tau)$ that combines states and rewards in the manners of DDR-I and DDR-II. 
% The DDR models are trained so that the noise can be well predicted by $\hat\epsilon_\theta\coloneqq\epsilon_\theta(\mathbf{x}_k(\tau), \mathbf{y}(\tau), k)$.
% Note that the conditional diffusion model and the IDM can be seen as separate training settings, 
% both using the samples directly from the dataset $\mathcal{D}$ for supervised training.

\vspace{-0.4em}
\subsection{Inverse Dynamics Models (IDMs)}
\label{sec:IDM}

Sampling states and rewards using a diffusion model is not enough for generating a full trajectory or extracting policy. 
A policy can be inferred from estimating the action $a_t$ that leads the state $s_t$ to $s_{t+1}$.
% at any time index $t$ in the trajectory $\tau$. 
\citet{Hepburn24} proposed implementing the IDM $p(a_t|s_t, s_{t+1})$ with a conditional variational autoencoder. 
However, they are generally harder to train with a more complicated structure than MLPs. 
We hence adopt the IDM from \citet{Pathak18},
which is also employed by DD and is shown to be effective than jointly diffusing over actions and states \citep{Ajay23}.
Let
%
%\begin{equation}
$    a_t\coloneqq f_\phi(s_t, s_{t+1})$,
%\end{equation}
%
where $f_\phi$ is an MLP with parameter $\phi$.
The training objective of IDM is simply
\vspace{-0.1em}
\begin{align}
    &\mathcal{L}_{\text{IDM}}(\phi)\coloneqq 
    \mathbb{E}_{(s, a, s')\in\mathcal{D}} 
    \ [\Vert a-f_\phi(s, s')\Vert^2 ].
\end{align}

\phantom{a}
\vspace{-3.4em}
\section{Stitcher of DITS}
\label{sec:TSKD}
\vspace{-0.3em}

As illustrated in Figure~\ref{fig:illustration},
merely generating high-return trajectories is sub-optimal as it may overfit those regions.
We overcome this limitation by blending such trajectories into the original dataset progressively with a \textit{selective} stitching strategy,
thereby diversifying the high-return trajectories by, \eg, 
starting from a low-reward region.

Stitching algorithms have been studied in the literature.
The Model-based Trajectory Stitching method \citep[MBTS,][]{Hepburn24} augments the dataset $\mathcal{D}$ by stitching together high-value regions of different trajectories,
\ie, introducing new transitions between two states on two different trajectories.
The candidate states for stitching are determined by 
i) the state-value function $V(s)$ and 
ii) the forward dynamics probability $p(s'|s)$. 
However, MBTS does not generate new trajectories conditioned with high return,
leaving the stitched data still sub-optimal.

% the optimal policy knowledge underlying in the dataset is not explicitly revealed by applying MBTS algorithm although the MBTS algorithm does help the BC algorithm to retrieve better policy knowledge.

To address this issue, 
we propose a stitching algorithm for DITS,
with an abridged version shown in Algorithm~\ref{alg:TSKD_skeleton},
and a detailed version deferred to Algorithm~\ref{alg:TSKD} in Appendix~\ref{sec:app_algo_details_TSKD}.
It can be compared with MBTS which is recapped as Algorithm~\ref{alg:MBTS} in Appendix~\ref{sec:app_algo_details_MBTS}.
The key differences are highlighted in color, 
and are summarized in Appendix~\ref{sec:app_algo_details_MBTS}.
% and the overall process is illustrated in Figure~\ref{fig:TSKD}.

% Our stitcher leverages DDRs to generate trajectories conditioned on high returns,
% and stitches with original trajectories to avoid overfitting high-return regimes by also covering other areas.
% Training on these TSKD-empowered trajectories by BC allows the DDR models' knowledge to be transferred to the learned policy,
% retaining similar performance to the deep models while significantly distilling them into a much shallower MLP.
% In this case, the DDR models are the teacher and the BC policy model is the student for knowledge distillation.

% TSKD leverages DDRs to retrieve the near optimal policy, 
% and to distill the policy knowledge via trajectory generation and stitching to transfer the knowledge into the BC model, 
% empowering it to have the performance close to the Decision Diffuser. 
% In this case, the DDR model is the teacher and the BC model is the student for the knowledge distilling setting. 

The key step of Algorithm~\ref{alg:TSKD_skeleton} is step \ref{alg_step:condition},
where a candidate state $\hat s'$ is evaluated for stitching.
Here, $\hat{.}$ stands for a candidate state.
By ``good'', we follow MBTS and impose three conditions as detailed in Algorithm~\ref{alg:TSKD}.
Firstly, it needs to be close to $s'$, 
and we detail the neighborhood selection in Appendix~\ref{sec:neighborhood}.
Moreover, there is a good chance to switch to $\hat s'$,
and the value of $\hat s'$ is higher than $s'$.
We will describe these two conditions in Section~\ref{sec:forward_dyn} and Appendix~\ref{sec:value_func}, respectively.

\begin{algorithm}[t]
\SetAlgoLined
   \caption{The stitching algorithm of DITS in brief. 
   A detailed version is in Algorithm~\ref{alg:TSKD}.}
   \label{alg:TSKD_skeleton}
   
   {\bfseries Input:} the original offline dataset $\mathcal{D}_0$ with $T$ trajectories $(\tau_1, ..., \tau_T)$
   
   \For{$k=0,...,K$}{
   
   Generate $n$ trajectories $\mathcal{D}'_{k}$ by {\color{blue} DITS' trajectory generator} in Algorithm~\ref{alg:DDR-I}.
   
   {\color{red} Set $\mathcal{D}_{k}\leftarrow \mathcal{D}_{k} \cup  \mathcal{D}'_{k}$}.
   Train the state-value function $V$ on dataset $\mathcal{D}_k$.
   
   \For{{\color{red} $t=1,...,T$ (i.e., length of $\mathcal{D}_k$)}} {
   
   $(s, s') \gets (s_0, s'_0)$ from $\tau_t$.  Initialize a new trajectory $\hat\tau_t$
   
   \While{not done (e.g., maximum length of episode not reached)}{
   \If{a state $\hat s'$ in $\mathcal{D}_k$ (possibly in a different trajectory than $t$) is \textbf{good} to stitch to \label{alg_step:condition}}{
   
   Generate action $\tilde a$ based on $(s, \hat s')$ by {\color{blue}DITS' action generator} (\eg, IDM).
   
   Generate reward $\tilde r$ given $(s, \hat s', \tilde a)$ by {\color{blue}DITS' reward generator} from Algorithm~\ref{alg:reward_gen}. \label{algo_label:reward_gen}
   
   Add $(s, \tilde a, \tilde r, \hat{s}')$ to the new trajectory $\hat \tau_t$
   
   $s \gets \hat s'$,   
   $s' \gets $ the next state on the trajectory of $\hat s'$
   }
   \Else{   
   Add original transition to $\hat \tau_t$ and slide $(s, s')$
   }
   }
   % \STATE \textbf{if} $\sum_{i \in \hat \tau_t}r_i \le (1 + \tilde p) \sum_{i \in \tau_t}r_i$}
   % \textbf{then} $\hat \tau_t \gets \tau_t$
   }

   {\color{red} $\mathcal{D}_{k+1} \gets T$ trajectories in $\{\hat \tau_t\}$ with the highest return}
   }
   {\bfseries Return:} dataset $\mathcal{D}_{K+1}$ for behavioural cloning training  
\end{algorithm}

\begin{algorithm}[t]
   \caption{Reward generation in DITS}
   \label{alg:reward_gen}
   
   {\bfseries Input:} Noise model $\epsilon_{\theta}$, guidance scale $\omega$, condition $\mathbf{y}$, current state $s$, stitching candidate $s'$
   
   Initialize $\mathbf{x}_K(\tau)\sim\mathcal{N}(\mathbf{0}, \alpha\mathbf{I})$ in $\mathbb{R}^{H(m+1)}$ from \Eqref{eq:xktau_def_ddr_1}
   
   \For{$k = K...1$}{
   
   $\mathbf{x}_k(\tau)[0].\text{state} \gets s$.
   \hspace{0.5em}
   $\mathbf{x}_k(\tau)[1].\text{state} \gets s'$.
   \hspace{0.5em}
   Set $\hat \epsilon$ with Equation \ref{classifier-free guidance}, using $\theta$, $\mathbf{x}_k(\tau)$, $\mathbf{y}$, and $k$.
   % $\hat{\epsilon}\leftarrow\epsilon_{\theta}(\mathbf{x}_k(\tau), \emptyset, k) + \omega(\epsilon_{\theta}(\mathbf{x}_k(\tau), \mathbf{y}, k) - \epsilon_{\theta}(\mathbf{x}_k(\tau), \emptyset, k))$
   %
     
   Sample $\mathbf{x}_{k-1}(\tau)\sim\mathcal{N}(\mu_{k-1}, \alpha\Sigma_{k-1})$,
   where $(\mu_{k-1}, \Sigma_{k-1})\leftarrow \text{Denoise}(\mathbf{x}_k(\tau), \hat\epsilon)$.
   
   }
   {\bfseries Return:} $\mathbf{x}_0(\tau)[0].\text{reward}$
   % , i.e., the $r_t$ in \eqref{eq:xktau_def_ddr_2}
\end{algorithm}

Whenever a stitching step occurs,
the following key query is required in step \ref{algo_label:reward_gen} of Algorithm \ref{alg:TSKD}:
\vspace{-0.4em}
\begin{quote}
    As we transition from state $s$ to state $\hat{s}'$,
    how much would the reward $r$ be?
\end{quote}

% This prediction will impact the decision of stitching/transitioning from trajectory $\tau_A$ (where $s$ lies) to $\tau_B$ (where $\hat s'$ lies). 
% In this case, it is necessary to model the reward based on both the current state and the next state. 
% Otherwise, the model will only predict the next reward in the trajectory $\tau_A$ while disregarding the information from $\tau_B$, 
% possibly leading to biased reward prediction for the stitching transition.

% So we develop two models for reward conditional diffusion:
% \begin{itemize}[leftmargin=*,topsep=0pt]
%    \setlength{\itemsep}{2pt}
%    \setlength{\parskip}{0pt}
% \item 
% \textbf{DDR-I}: a diffuser that directly generates full trajectories.
% The reward only needs to depend on the most recent history of states.
% %
% \item 
% \textbf{DDR-II}: a diffuser that generates reward depending on the current state \textit{and} the next proposed state, without action.
% \end{itemize}

\vspace{-0.5em}
Since the state transition $s \to \hat{s}'$ is not present in the data,
we need to complement it with reward.
A straightforward solution is to reuse the trajectory diffuser by clamping the recent two states to $s$ and $\shat'$.
The details are given in Algorithm~\ref{alg:reward_gen}.
We also tried to train a separate diffusion model for this purpose, and got similar results.
The details are deferred to Appendix \ref{Sec:app_gen_reward}.

\subsection{Forward Dynamics Criterion}
\label{sec:forward_dyn}

To determine whether a state can be a candidate next state for stitching, 
we need a criterion based on forward dynamics model $p(\hat s'|s)$, 
depicting the transition probability from state $s$ to the possible candidate state $\hat s'$, 
which may \textit{differ} from the observed next state $s'$. We model the environment dynamics as a Gaussian distribution, which is common for continuous state-space applications \citep{Janner19}. 
We create an ensemble $\mathcal{E}$ of $N$ dynamics models parameterized by $\xi^i$:
$\{\hat p_{\xi^i}(s_{t+1}|s_t)=\mathcal{N}(\mu_{\xi^i}(s_t), \Sigma_{\xi^i}(s_t))\}^N_{i=1}$. 
Each model is trained via maximum likelihood estimation from the dataset $\mathcal{D}$, for which the loss can be formulated as:
\begin{align}
    &\mathcal{L}_{\hat p}(\xi)\coloneqq  \mathbb{E}_{s, s'\sim\mathcal{D}} [(\mu_\xi(s) - s')^T\Sigma_\xi^{-1}(s)(\mu_\xi(s)-s')+\log\vert\Sigma_\xi(s)\vert ],
\end{align}
where $\vert\cdot\vert$ is the determinant of a matrix. We train the ensemble using different parameter initializations for each model in order to take the epistemic uncertainty into account.
The criterion for determining the candidate next state can be formulated in a conservative manner:
\begin{align}
    \min\nolimits_{i\in\mathcal{E}}
    % \min\nolimits_{i\in\mathcal{E}}
    \ \hat p_{\xi^i}(\hat s'|s) > 
    \mathop{\text{mean}}\nolimits_{i\in\mathcal{E}}
    % \mathop{\text{mean}}\nolimits_{i\in\mathcal{E}} 
    \ \hat p_{\xi^i}(s'|s).
\end{align}

\begin{table*}
\caption{Comparison of \textbf{D4RL score} between DITS and  \textbf{interaction-free} augmentation methods,
including SynthER and MBTS.
In general, our DITS shows significant improvement.
The base offline RL methods are BC and TD3+BC.
We ran 5 random seeds.}
\label{result-table}
\vspace{-0.2em}
\setlength\tabcolsep{3pt}
% \vskip 0.15in
\begin{center}
\begin{small}
\begin{tabular}{l|cccc|cccc}
\toprule
    & \multicolumn{4}{c|}{BC} & \multicolumn{4}{c}{TD3+BC} \\
  \midrule
Methods & Original&SynthER&MBTS&\textbf{DITS}&Original&SynthER&MBTS&\textbf{DITS} \\
\midrule
&\multicolumn{8}{c}{\textbf{Med-Expert}}\\
\midrule
% Dataset & Environment &  &  &  &  &  &  &  & \\
% \midrule
Halfcheetah & 55.2{\scriptsize $\pm$8.3} & 61.9{\scriptsize $\pm$7.8} & \textbf{86.9{\scriptsize $\pm$2.5}} & 86.8{\scriptsize $\pm$0.7} & 90.8{\scriptsize $\pm$7.0} & 85.9{\scriptsize $\pm$8.2} & \textbf{93.8{\scriptsize $\pm$3.4}} & 91.2{\scriptsize $\pm$3.7}\\
Hopper& 52.5{\scriptsize $\pm$7.7} & 61.2{\scriptsize $\pm$8.8} & 94.8{\scriptsize $\pm$11.7} & \textbf{105.1{\scriptsize $\pm$6.1}} & 101.1{\scriptsize $\pm$10.5} & 102.5{\scriptsize $\pm$10.9} & 109.1{\scriptsize $\pm$3.9} & \textbf{111.3{\scriptsize $\pm$3.6}} \\
Walker2d & 107.5{\scriptsize $\pm$2.8} & 108.2{\scriptsize $\pm$3.3} & 108.8{\scriptsize $\pm$5.5} & \textbf{109.1{\scriptsize $\pm$0.1}} & 110.0{\scriptsize $\pm$0.4} & 110.1{\scriptsize $\pm$0.3} & 110.3{\scriptsize $\pm$0.4} & \textbf{110.4{\scriptsize $\pm$0.8}}  \\
\midrule
\textbf{Average} & 71.7{\scriptsize $\pm$6.3} & 77.1{\scriptsize $\pm$6.7} & 96.8{\scriptsize $\pm$6.6} & \textbf{100.3{\scriptsize $\pm$2.3}} & 100.6{\scriptsize $\pm$6.0} & 99.5{\scriptsize $\pm$6.5} & \textbf{104.4{\scriptsize $\pm$2.6}} & 104.3{\scriptsize $\pm$2.7}\\
\midrule
&\multicolumn{8}{c}{\textbf{Medium}}\\
\midrule
Halfcheetah & 42.6{\scriptsize $\pm$3.1} & 42.8{\scriptsize $\pm$2.3} & 43.2{\scriptsize $\pm$0.3} & \textbf{44.1{\scriptsize $\pm$0.7}} & 48.1{\scriptsize $\pm$0.2} & 48.8{\scriptsize $\pm$0.3} & 48.4{\scriptsize $\pm$0.4} & \textbf{50.3{\scriptsize $\pm$3.4}} \\
Hopper & 52.9{\scriptsize $\pm$2.4} & 58.1{\scriptsize $\pm$5.7} & 64.3{\scriptsize $\pm$4.2} & \textbf{83.2{\scriptsize $\pm$8.1}} & 60.4{\scriptsize $\pm$4.0} & 63.0{\scriptsize $\pm$4.3} & 64.1{\scriptsize $\pm$4.4} & \textbf{76.6{\scriptsize $\pm$6.4}} \\
Walker2d & 75.3{\scriptsize $\pm$6.2} & 74.7{\scriptsize $\pm$5.5} & \textbf{78.8{\scriptsize $\pm$1.2}} & 78.5{\scriptsize $\pm$2.0} & 82.7{\scriptsize $\pm$5.5} & \textbf{85.2{\scriptsize $\pm$1.1}} & 84.2{\scriptsize $\pm$1.4} & 83.3{\scriptsize $\pm$3.6} \\
\midrule
\textbf{Average} & 56.9{\scriptsize $\pm$3.9} & 58.5{\scriptsize $\pm$4.5} & 62.1{\scriptsize $\pm$1.9} & \textbf{68.6{\scriptsize $\pm$3.6}} & 63.7{\scriptsize $\pm$3.2} & 65.7{\scriptsize $\pm$1.9} & 65.6{\scriptsize $\pm$2.1} & \textbf{70.2{\scriptsize $\pm$4.5}} \\
\midrule
&\multicolumn{8}{c}{\textbf{Med-Replay}}\\
\midrule
Halfcheetah & 36.6 {\scriptsize $\pm$3.1}& 34.1{\scriptsize $\pm$2.5} & \textbf{39.8{\scriptsize $\pm$0.6}} & 38.9{\scriptsize $\pm$3.2} & 44.6{\scriptsize $\pm$3.3} & 44.7{\scriptsize $\pm$0.6} & 43.8{\scriptsize $\pm$0.5} & \textbf{45.1{\scriptsize $\pm$5.5}}\\
Hopper & 18.1{\scriptsize $\pm$7.4} & 27.5{\scriptsize $\pm$10.2} & 50.2{\scriptsize $\pm$17.2} & \textbf{96.4{\scriptsize $\pm$6.8}} & 60.9{\scriptsize $\pm$5.2} & 63.2{\scriptsize $\pm$4.4} & 77.4{\scriptsize $\pm$17.0} & \textbf{99.4{\scriptsize $\pm$1.9}} \\
Walker2d & 26.0{\scriptsize $\pm$5.9} & 33.1{\scriptsize $\pm$10.8} & 61.5{\scriptsize $\pm$5.6} & \textbf{74.5{\scriptsize $\pm$6.5}} & 81.8{\scriptsize $\pm$5.3} & 82.1{\scriptsize $\pm$5.6} & 82.8{\scriptsize $\pm$3.4} & \textbf{83.4{\scriptsize $\pm$4.2}} \\
\midrule
\textbf{Average}& 26.9{\scriptsize $\pm$5.5} & 31.6{\scriptsize $\pm$7.8} & 50.5{\scriptsize $\pm$7.8} & \textbf{69.9{\scriptsize $\pm$5.5}} & 62.4{\scriptsize $\pm$4.6} & 63.3{\scriptsize $\pm$3.5} & 68.0{\scriptsize $\pm$7.0} & \textbf{76.0{\scriptsize $\pm$3.9}} \\
\midrule
&\multicolumn{8}{c}{\textbf{Kitchen}}\\
\midrule
Mixed & 51.5{\scriptsize $\pm$7.3} & 53.7{\scriptsize $\pm$5.8} & 49.3{\scriptsize $\pm$6.6} & \textbf{60.0{\scriptsize $\pm$4.8}} & 0.0{\scriptsize $\pm$0.0} & 0.0{\scriptsize $\pm$0.0} & 4.3{\scriptsize $\pm$2.2} & \textbf{42.2{\scriptsize $\pm$6.3}} \\
Partial & 38.0{\scriptsize $\pm$5.3} & 44.2{\scriptsize $\pm$5.5} & 39.4{\scriptsize $\pm$7.8} & \textbf{44.4{\scriptsize $\pm$3.9}} & 0.7{\scriptsize $\pm$0.4} & 0.5{\scriptsize $\pm$0.2} & 2.5{\scriptsize $\pm$0.8} & \textbf{37.4{\scriptsize $\pm$7.1}} \\
\midrule
\textbf{Average} & 44.8{\scriptsize $\pm$6.3} & 49.0{\scriptsize $\pm$5.7} & 44.4{\scriptsize $\pm$7.2} & \textbf{52.2{\scriptsize $\pm$4.4}} & 0.4{\scriptsize $\pm$0.2} & 0.3{\scriptsize $\pm$0.1} & 3.4{\scriptsize $\pm$1.5} & \textbf{39.8{\scriptsize $\pm$6.7}} \\
\midrule
&\multicolumn{8}{c}{\textbf{Antmaze}}\\
\midrule
Umaze & 55.3{\scriptsize $\pm$4.2} & 66.5{\scriptsize $\pm$4.4} & 73.6{\scriptsize $\pm$7.3} & \textbf{82.3{\scriptsize $\pm$10.2}} & 70.8{\scriptsize $\pm$39.2}& 87.3{\scriptsize $\pm$6.6} & 75.6{\scriptsize $\pm$13.8} & \textbf{92.3{\scriptsize $\pm$7.7}}\\
U-Diverse & 47.3{\scriptsize $\pm$4.1} & 58.4{\scriptsize $\pm$3.8} & 63.8{\scriptsize $\pm$4.8} & \textbf{65.5{\scriptsize $\pm$8.8}}& 44.8{\scriptsize $\pm$11.6} & 57.8{\scriptsize $\pm$7.6} & 59.7{\scriptsize $\pm$6.3} & \textbf{65.2{\scriptsize $\pm$6.3}}\\
\midrule
\textbf{Average} & 51.3{\scriptsize $\pm$4.2} & 62.5{\scriptsize $\pm$4.1} & 68.7{\scriptsize $\pm$6.1} & \textbf{73.9{\scriptsize $\pm$9.5}} & 57.8{\scriptsize $\pm$25.4} & 72.6{\scriptsize $\pm$7.1} & 67.7{\scriptsize $\pm$10.1} & \textbf{78.8{\scriptsize $\pm$7.0}} \\
\bottomrule
\end{tabular}
\end{small}
\end{center}
\vskip -0.16in
\end{table*}

\begin{table}[]
    \centering
    \caption{
    Comparison of \textbf{D4RL score} between DITS and  \textbf{interaction-based} augmentation methods,
    including the vanilla SynthER and its modified version.
In general, our DITS shows significant improvement.
The base offline RL method is \textbf{TD3+BC}.
We ran 5 random seeds.}
    \begin{tabular}{lc|ccr}
        \toprule
        % \multicolumn{2}{c|}{} & \multicolumn{3}{c}{TD3+BC}\\
        % \midrule
       \multicolumn{2}{c|}{} & SynthER (original) & SynthER (modified) & DITS  \\
       \midrule
       Med-Expert & Halfcheetah & 85.9$\pm$8.2 & 88.3$\pm$4.1 & \textbf{91.2$\pm$3.7}\\
       Med-Expert & Hopper & 102.5$\pm$10.9 & 101.9$\pm$5.7& \textbf{111.3$\pm$3.6}\\
       Med-Expert & Walker2d & 110.1$\pm$0.3 & 110.2$\pm$0.5&\textbf{110.4$\pm$0.8}\\
       \midrule
        Medium & Halfcheetah& 48.4$\pm$0.3 & 47.3$\pm$1.5&\textbf{50.3$\pm$3.4}\\
       Medium & Hopper & 63.0$\pm$4.3& 65.8$\pm$2.9&\textbf{76.6$\pm$6.4}\\
       Medium & Walker2d& 85.2$\pm$1.1 & \textbf{85.4$\pm$2.8}&83.3$\pm$3.6\\
       \midrule
       Med-Replay & Halfcheetah & 44.7$\pm$0.6& 44.0$\pm$0.9 & \textbf{45.1$\pm$5.5}\\
       Med-Replay & Hopper & 63.2$\pm$4.4 & 68.5$\pm$2.8 & \textbf{99.4$\pm$1.9}\\
       Med-Replay & Walker2d & 82.1$\pm$5.6 & \textbf{84.1$\pm$5.3} & 83.4$\pm$4.2\\
       \bottomrule
    \end{tabular}
    \label{modified-test}
\vspace{-0.4em}        
\end{table}

% \begin{itemize}
% \item
% \textbf{Task oriented} 
% There are two different tasks for the conditional diffusion model to complete. For the first one, the model is required to generate full trajectories, which for this case modeling the reward only depending on the current state $s_t$ is sufficient. 
% For the second one, however, during the trajectory stitching process, 
% the model is required to generate reward based on the current state $s$ and the selected next state $\hat{s}'$ to perform a stitching from trajectory $\tau_A$ to $\tau_B$. 
% In this case, modeling the reward to depend on both the current state and the next state is necessary. Otherwise, the model will only predict the next reward in the trajectory $\tau_A$ disregarding the information from $\tau_B$, which may possibly lead to biased reward prediction for the stitching transition.

% \item\textbf{Limitations in the phase of trajectory generation} During the online generation phase, we can only observe one state $s_0$ initially from the environment, which prevents us from using the model that requires the current state $s_0$ and the next state $s_1$ for the diffusion. Therefore, we choose to use DDR-I model, which doesn't need the observation for the next state during evaluation, for trajectory generation task, while using the DDR-II model for the reward generation task.
% \end{itemize}

\vspace{-0.9em}
\section{Experiments}
\label{sec:exp}
\vspace{-0.2em}

We now demonstrate the empirical performance of DITS for offline RL with state-only interactions.
We emphasize that our goal is data augmentation, 
instead of proposing a new RL algorithm that possibly  intertwines with or extends another existing RL algorithm delicately.
Therefore, the comparison is against other augmentation methods under a common set of base RL algorithms.

\vspace{-0.5em}
\paragraph{Environments} 

Although the three applications listed in the introduction are realistic in practice,
we do not have access to their offline data.
As a workaround,
we employed standard locomotion tasks from the D4RL benchmark \citep{Fu20},
including Halfcheetah, Hopper, and Walker2d. 
To explore its robustness, 
we further tested on Kitchen and Antmaze, 
two harder environments. 
We only considered the medium-expert, medium, and medium-replay datasets for locomotion tasks,
because the downstream RL models already have excellent performance in expert datasets.

\paragraph{Baseline data augmentation methods}

We first compared with state-of-the-art interaction-free data augmentation methods,
including SynthER \citep{lu2023synthetic} and MBTS \citep{Hepburn24}.
Then we compared with an extended version of SynthER that utilizes state-only interactions.
In particular,
as SynthER generates $(s, a, r, s')$,
we replaced the diffusion generated next state $s'$ with the actual next state retrieved from interaction. 
We kept the same number of generated samples between modified SynthER and DITS.

\vspace{-0.5em}
\paragraph{Settings}
We evaluated the performance using the \textit{normalized average return} \citep{Fu20}.
It first applies the offline learned policy to 10 online episodes.
The total reward of each episode is referred to as a \textit{score},
which is then normalized via $100 \times \frac{\text{score}-\text{random score}}{\text{expert score}-\text{random score}}$.
The process is repeated with 5 random seeds,
producing mean and variance.
DITS used $300$ trajectories generated from the DITS trajectory generator. 
The detailed hyperparameter settings are relegated to Appendix~\ref{hyperparams}.

\iffalse
% Our method shows superiority on the evaluation scores over a variety of tasks. 
On the one hand, for the comparisons with interaction-free methods, we train BC and TD3+BC model on the augmented datasets. 
On the other hand, we only compare with the modified SynthER on a trained TD3+BC model over the augmented dataset. 
Since BC model doesn't utilize the next state $s'$, it will show no difference between the modified SynthER and the original version.

\vspace{-0.5em}
\paragraph{SynthER with Interaction} 
In order to form a fair comparison, we introduce a modified version of SynthER that includes interactions with the environment. We replace the diffusion-generated next state $\tilde s'$ with the next state $s'$ from actual interactions with the environment, given the current state $s$ and action $a$. We keep the number of generated samples the same between modified SynthER and our DITS for the comparison results in Table~\ref{modified-test}.
\fi

\vspace{-0.4em}    
\subsection{Performance of normalized average return}

The results of normalized average score for comparisons with \textit{interaction-free} methods are shown in Table~\ref{result-table},
where the base RL learners are behavioral cloning (BC) and TD3+BC \citep{fujimoto2021minimalist}.
A similar comparison with \textit{interaction-based} methods is shown in Table~\ref{modified-test},
where we only compared with the modified SynthER on TD3+BC,
because BC does not utilize the next state $s'$,

Our DITS  made significant improvement on the Kitchen tasks for TD3+BC. 
The original TD3+BC performs poorly in this situation, 
along with the other augmentation methods under comparison. 
However, DITS is able to generate full demonstrative trajectories with good policies using conditional diffusion based on state-only interactions, 
resulting in a dramatic boost for the TD3+BC's performance. 
The improvement offered by DITS is more significant when  the original dataset is not of high quality.
Comparing the two columns of SynthER in Table~\ref{modified-test},
it does get improved by state-only interactions.

For a broader range of base RL methods,
we evaluated our DITS augmentation based on CQL \citep{kumar2020conservative}, 
IQL \citep{Kostrikov21}, 
and CPED \citep{zhang2024constrainedpolicyoptimizationexplicit}. 
% We observed performance improvement on locomotion datasets. 
Table \ref{tab:merged-results} presents the performance on Hopper and Walker2d.
Our DITS is generally able to improve the performance,
especially on datasets with lower quality.

\vspace{-0.5em}
\paragraph{Trajectory generation time cost}
We evaluated the time consumption of  trajectory generation for DITS.
Using Hopper Medium Expert as an example,
the mean and standard deviation in seconds over 10 repeated trials are:
236.89$\pm$5.93 (200 steps),
474.69$\pm$11.24 (400 steps), 711.63$\pm$17.08 (600 steps), 950.03$\pm$21.85 (800 steps), 1185.19$\pm$23.83 (1000 steps).
We ran on a single RTX3080 GPU.

% \begin{table}[ht]
% \label{time-table}
% \setlength\tabcolsep{3pt}
% \vskip 0.15in
% \begin{center}
% \begin{small}
% \begin{tabular}{lccccr}
% \toprule 
%  & 200 steps & 400 steps & 600 steps & 800 steps & 1000 steps\\
% \midrule
% Generation time consumption& 236.89$\pm$5.93 & 474.69$\pm$11.24 & 711.63$\pm$17.08 & 950.03$\pm$21.85 & 1185.19$\pm$23.83\\
% \bottomrule
% \end{tabular}
% \end{small}
% \end{center}
% \vskip -0.1in
% \caption{\textbf{Time Consumption during Trajectory Generation} We evaluated the time consumption during trajectory generation phase in Hopper Medium Expert. We computed the average and standard deviation using 10 repeated experiments.}
% \end{table}

\begin{table}[t]
\caption{Improvement of D4RL score as DITS is applied in conjunction with CQL, IQL, and CPED.
% \color{red}{improve the performance of CPED agent in some of the scenarios
% }
}
\label{tab:merged-results}
\begin{center}
\begin{small}
\begin{tabular}{lc|cc|cc|cc}
\toprule
\multicolumn{2}{r|}{} & CQL & DITS+CQL & IQL & DITS+IQL & CPED & DITS+CPED \\
\midrule
Hopper& Medium-Expert & 105.4 & \textbf{109.3$\pm$2.8} & 91.5 & \textbf{111.2$\pm$2.1} & 95.3$\pm$13.5 & \textbf{109.4$\pm$2.1} \\
Hopper& Medium & 58.5 & \textbf{77.2$\pm$7.5} & 66.3 & \textbf{77.9$\pm$5.5} & \textbf{100.1$\pm$2.8} & 97.9$\pm$4.9 \\
Hopper& Medium-Replay & 95.0 & \textbf{96.8$\pm$4.7} & 94.7 & \textbf{101.9$\pm$1.3} & 98.1$\pm$2.1 & \textbf{99.3$\pm$1.8} \\
\midrule
Walker2d& Medium-Expert & \textbf{108.8} & \textbf{108.3$\pm$1.5} & 108.8 & \textbf{111.1$\pm$1.3} & \textbf{113.0$\pm$1.4} & 110.8$\pm$2.2 \\
Walker2d& Medium & 72.5 & \textbf{79.1$\pm$2.9} & 78.3 & \textbf{84.1$\pm$3.3} & 90.2$\pm$1.7 & \textbf{93.4$\pm$2.0} \\
Walker2d& Medium-Replay & 77.2 & \textbf{79.0$\pm$3.8} & 73.9 & \textbf{87.9$\pm$4.1} & 91.9$\pm$0.9 & \textbf{92.1$\pm$1.1} \\
\bottomrule
\end{tabular}
\end{small}
\end{center}
\end{table}

\begin{table}[t]
    \caption{\textbf{Correlation analysis.} 
    While SynthER reconstructs the data from the original training set, we show that our method can improve the quality of the data with a higher similarity to the expert data. We used 100k samples from each model to compute the average result.}
    \label{matching-test}
    \centering
    \begin{tabular}{l|cc|cc}
        \toprule
        & \multicolumn{2}{c|}{SynthER} & \multicolumn{2}{c}{DITS}  \\
       % \midrule
        & Marginal & Correlation & Marginal & Correlation\\
       \midrule
       Hopper Med-Replay & \textbf{0.983} & \textbf{0.997} & 0.932 & 0.989\\
       Hopper Med-Expert & 0.958 & 0.992 & \textbf{0.982} & \textbf{0.995}\\
       Hopper Expert & 0.934 & 0.982 & \textbf{0.998} & \textbf{0.998}\\
       \bottomrule
    \end{tabular}
\vspace{-0.9em}    
\end{table}

\begin{figure}[t]
\centering
\begin{subfigure}{0.325\textwidth}
  \centering  \includegraphics[width=\linewidth]{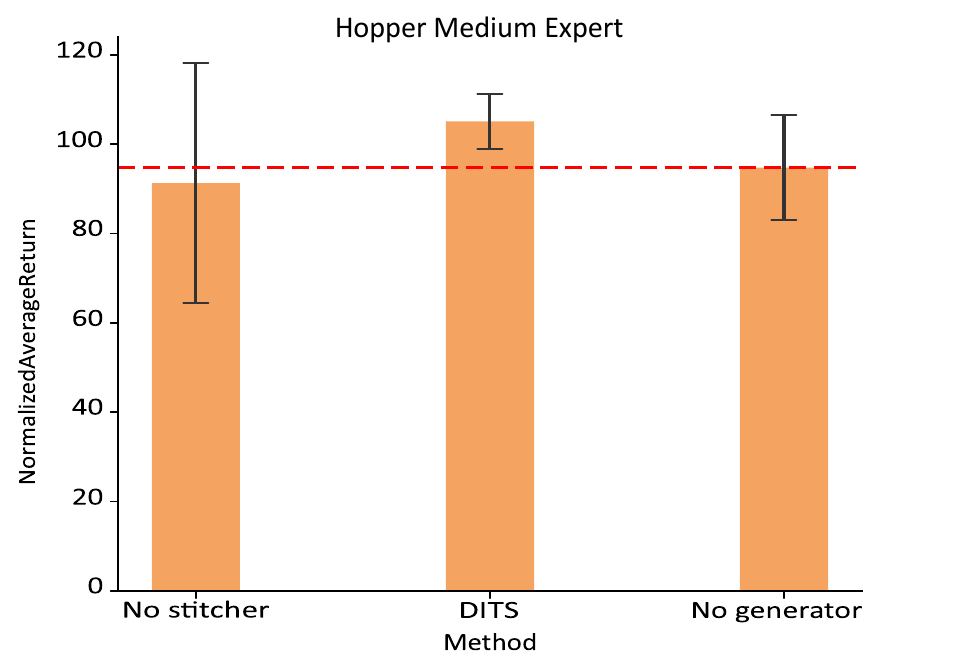}\quad
  \caption{medium expert}
  \label{fig:1_medium_expert}
\end{subfigure}
\begin{subfigure}{0.325\textwidth}
  \centering  \includegraphics[width=\linewidth]{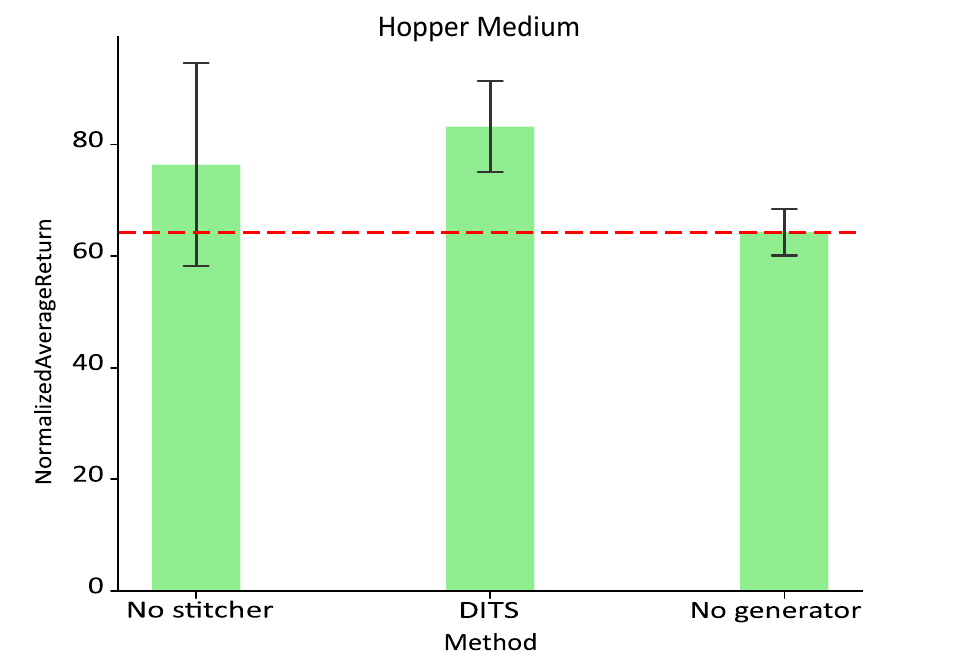}\quad
  \caption{medium}
  \label{fig:1_medium}
\end{subfigure}
\begin{subfigure}{0.325\textwidth}
  \centering  \includegraphics[width=\linewidth]{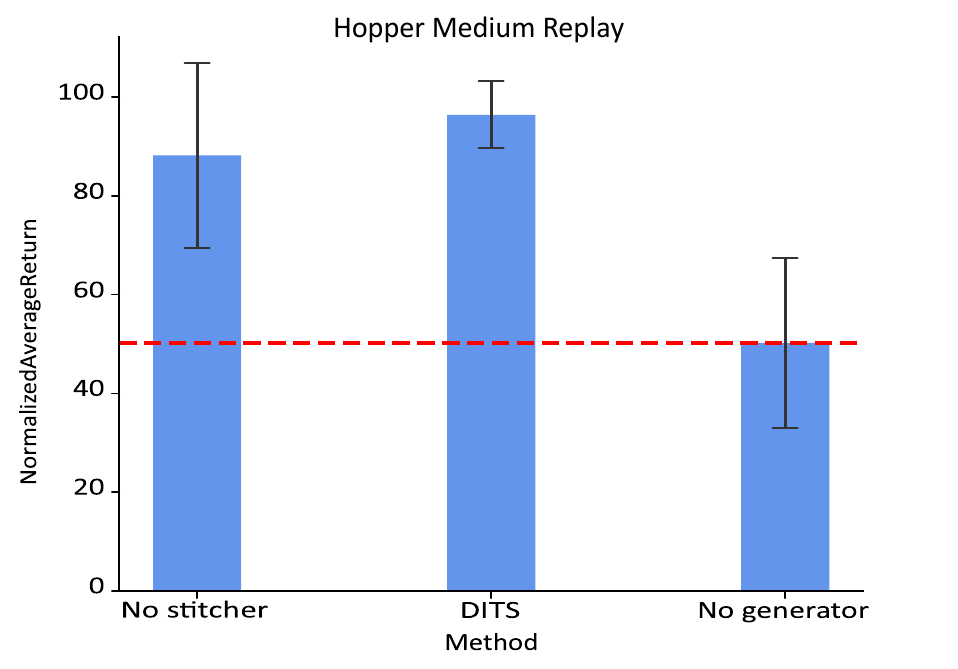}
  \caption{medium replay}
  \label{fig:1_medium_replay}
\end{subfigure}
\vspace{-0.3em}
\caption{\textbf{Ablation study on trajectory generation and stitching}. 
We show the contribution of the two components of DITS by comparing its \textit{normalized average return} on Hopper with two variants: no stitcher (dropping the stitching step) and 
no generator (dotted line, using the original dataset).
% (it's typically the original MBTS method). 
% We used the Hopper environment,
% and the base RL algorithm is BC.
% a) dropping the stitching step (DDR+BC); 
% and b) dropping DDR-I (generating high-return trajectories) and DDR-II (which is required by TSKD). which is approximated by MBTS+BC.
}
\label{ablation-hopper}
% \vspace{-1em}
\end{figure}

% \vspace{-0.5em}
\subsection{Correlation Analysis}
\label{Correlation-DITS}

Following \citet{lu2023synthetic},
we next measured the similarity between 
a) transitions generated SynthER and DITS diffusion model (without stitching) after training from the Hopper \textit{Medium Replay} dataset,
and 
b) samples from the \textit{Medium Replay}, \textit{Medium Expert} and \textit{Expert} datasets.
The metrics included the Kolmogorov-Smirnov statistic and the Pearson rank correlation \citep{patki2016synthetic}. 
% We trained our conditional diffusion model and the SynthER diffusion model on the Hopper Medium Replay dataset and evaluated the results with data retrieved from Medium Replay, Medium Expert and Expert datasets. 
As shown in Table~\ref{matching-test},
DITS can generate data with the distribution close to the expert dataset while training only on medium replay data thanks to the conditional guidance. 
We refer to the KS test results as \textbf{Marginal},
and the Pearson statistics results as \textbf{Correlation}.

\vspace{-0.5em}
\subsection{Ablation Study on the DITS Method}
\label{ablation_TSKD}

Our first ablation study investigates how the two key ingredients of our method,
the trajectory generator and the stitcher, 
contribute to its performance.
Using BC as the base RL algorithm,
We compared DITS with two variants:
a) only adding DITS generated trajectories to the datasets based on state-only interactions,
and 
b) only performing stitching on the original dataset without generating new trajectories.

The resulting normalized average score is presented in Figure~\ref{ablation-hopper},
using the Hopper dataset and the same three difficulty levels as in Table~\ref{result-table}. 
The results for Halfcheetah and Walker2d are deferred to Figure~\ref{ablation-addition} in Appendix~\ref{sec:ablation_DITS}.
Overall, fully applying DITS is more effective than only utilizing the stitcher, 
which indicates that the conditional diffuser and interaction are indeed helpful by providing high-return trajectories.
Furthermore, full DITS outperforms the method with stitcher removed,
because the stitching step can avoid overfitting to the high-return regions by blending in transitions from the low value regions.

\vspace{-0.5em}
\paragraph{Ablation on Conditional Guidance Scale}
We also analyzed the sensitivity of the conditional guidance scale $\omega$ of the conditional diffusion model.
On the Hopper Medium Expert dataset,
we evaluated the generation quality by computing the correlation between expert data. 
The correlation turns out 
$0.996, 0.996, 0.993, 0.989$
for $\omega = 1.2, 1.4, 1.6, 1.8$, respectively.

\vspace{-0.5em}
\paragraph{Ablation on the Number of Generated Trajectories}
% \label{ablation_traj_num}

We next study the impact of the high-return trajectories generated by DITS via interaction.
Figure~\ref{ablation-DDR} shows that, using Hopper medium expert as the environment and BC as the base offline RL algorithm,
the performance of our method can be improved with more sampled trajectories.
Further, the variance also decays as a broader range of situations are fed to BC.
When only a small number of high-return trajectories are available,
some test episodes turned out to suffer a considerably low score.

\subsection{Comparison with DiffStitch}
\label{sec:compare_diffstitch_app}

DiffStitch \citep{li2024DiffStitch} is a recent work that uses diffusion imaginations to stitch trajectories that augment the data. 
However, it does not utilize state-only interactions,
nor employ conditional guidance. 
Although we also use diffusion models,
we use them to generate high-reward trajectories as candidates for stitching,
instead of using them to implement the stitching method itself. 

\begin{figure}
\begin{minipage}[b]{0.4\textwidth}
\centerline{\includegraphics[width=\textwidth]{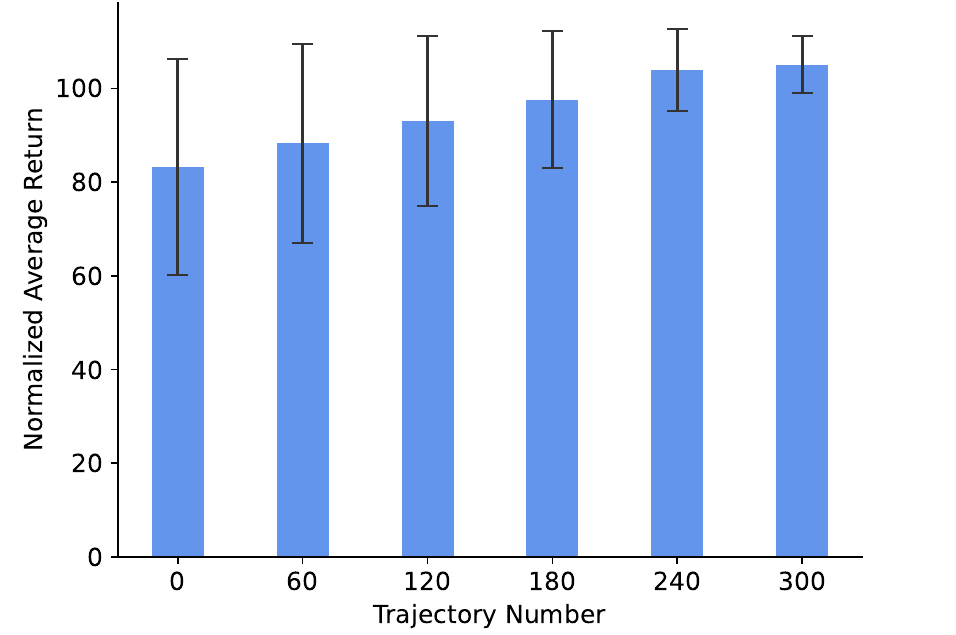}}
\vspace{-0.8em}
\caption{Ablation on the number of DITS generated trajectory 
% As we gradually reduce the number of generated trajectories from DITS trajectory generator that are fed into the trajectory stitching process,
% the performance degrades significantly. We evaluate the ablation on BC policy.
% Data: Hopper medium expert.
}
\label{ablation-DDR}
\end{minipage}
~~~
\begin{minipage}[b]{0.55\textwidth}
    \centering
% \begin{table}[h]
    \centering
    \begin{tabular}{lccr}
        \toprule
        % \multicolumn{2}{c}{} & \multicolumn{2}{c}{TD3+BC}\\
        % \midrule
       \multicolumn{2}{c}{} & DiffStitch & DITS  \\
       \midrule
       Med-Expert & Halfcheetah & \textbf{96.0$\pm$0.5} & 91.2$\pm$3.7\\
       Med-Expert & Hopper & 107.1$\pm$7.0& \textbf{111.3$\pm$3.6}\\
       Med-Expert & Walker2d & \textbf{110.2$\pm$0.3}& \textbf{110.4$\pm$0.8} \\
       \midrule
        Medium & Halfcheetah& \textbf{50.4$\pm$0.5} &\textbf{50.3$\pm$3.4}\\
       Medium & Hopper & 60.3$\pm$4.9&\textbf{76.6$\pm$6.4}\\
       Medium & Walker2d& \textbf{83.4$\pm$4.7} &\textbf{83.3$\pm$3.6}\\
       \midrule
       Med-Replay & Halfcheetah & 44.7$\pm$0.3 & \textbf{45.1$\pm$5.5}\\
       Med-Replay & Hopper & 79.6$\pm$13.5 & \textbf{99.4$\pm$1.9}\\
       Med-Replay & Walker2d & \textbf{89.7$\pm$4.2} & 83.4$\pm$4.2\\
       \bottomrule
    \end{tabular}
    \captionof{table}{Comparing DITS and DiffStitch in D4RL score}
    \label{result-diffstitch}
\end{minipage}
\end{figure}

For completeness we compare in Table~\ref{result-diffstitch} the D4RL scores of our DITS and DiffStitch. 
The base RL method is TD3+BC,
and 5 random seeds were used.
Our DITS matches or outperforms DiffStitch in D4RL locomotion datasets.

Due to space limitation,
we defer to Appendix~\ref{sec:reward_assess} the accuracy of reward generation.
% b) comparison with DiffStitch \citep{li2024DiffStitch} as in Appendix~\ref{sec:compare_diffstitch_app}.

\vspace{-0.5em}
\section{Conclusion, Limitation, Broader Impact, and Future work}

We proposed a novel data augmentation method DITS for offline RL,
where state-only interactions are available with the environment.
The generator based on conditional diffusion models allows high-return trajectories to be sampled,
and the stitching algorithm blends them with the original ones.
The resulting augmented dataset is shown to significantly boost the performance of base RL methods.

% TSKD can be extended to multi-teacher distillation \citep{You17} by training a set of DDR models from different knowledge sources. 
% Besides combining their generated trajectories,
% different DDR-II models can also improve TSKD by providing epistemic uncertainty of reward generation.

As a limitation, we did not update DITS' trajectory generator after new state-only sequences are obtained from the interaction.
Future work will empower diffusion models to learn from partially observed data (no reward available).
As for broader impact, DITS will benefit a range of applications where state-only interactions are available,
\eg, healthcare, recommendation, and inventory control.

\bibliography{iclr2025_conference}
\bibliographystyle{iclr2025_conference}

%%%%%%%%%%%%%%%%%%%%%%%%%%%%%%%%%%%%%%%%%%%%%%%%%%%%%%%%%%%%%%%%%%%%%%%%%%%%%%%
%%%%%%%%%%%%%%%%%%%%%%%%%%%%%%%%%%%%%%%%%%%%%%%%%%%%%%%%%%%%%%%%%%%%%%%%%%%%%%%
% APPENDIX
%%%%%%%%%%%%%%%%%%%%%%%%%%%%%%%%%%%%%%%%%%%%%%%%%%%%%%%%%%%%%%%%%%%%%%%%%%%%%%%
%%%%%%%%%%%%%%%%%%%%%%%%%%%%%%%%%%%%%%%%%%%%%%%%%%%%%%%%%%%%%%%%%%%%%%%%%%%%%%%
\newpage
\appendix

% \begin{center}
%     \Large Supplementary Material
% \end{center}

\section{Algorithm Details}
\label{sec:app_algo_details}

\subsection{Diffusion-based Trajectory Stitching (DITS)}
\label{sec:app_algo_details_TSKD}

We present in Algorithm~\ref{alg:TSKD} the detailed pseudo-code for the stitching algorithm of DITS.

\iffalse
\begin{algorithm}[H]
   \caption{Reward Generation with DDR-II}
   \label{alg:DDR-II}
\begin{algorithmic}[1]
   \STATE {\bfseries Input:} Noise model $\epsilon_{\theta'}$, guidance scale $\omega$, condition $\mathbf{y}$, current state $s$, stitching candidate $s'$
   \STATE $B\leftarrow (s, s', \emptyset)$
   \STATE $input.insert(B)$
   \STATE Initialize $\mathbf{x}_K(\tau)\sim\mathcal{N}(\mathbf{0}, \alpha\mathbf{I})$
   \FOR{$k = K...1$}   
   \IF{$k=K$}
   \STATE $input_{cond} \leftarrow input$
   \ELSE
   \STATE Extract $r_k$ from $x_k(\tau)$
   \STATE Replace all $r$ in $input_{cond}$ with $r_k\texttt{[}:length(input_{cond})\texttt{]}$
   \ENDIF
   \STATE 
   $\mathbf{x}_k(\tau)\leftarrow B_{cond}$
   \STATE$\hat{\epsilon}\leftarrow\epsilon_{\theta'}(\mathbf{x}_k(\tau), \emptyset, k) + \omega(\epsilon_{\theta'}(\mathbf{x}_k(\tau), \mathbf{y}, k) - \epsilon_{\theta'}(\mathbf{x}_k(\tau), \emptyset, k))$
   \STATE $(\mu_{k-1}, \Sigma_{k-1})\leftarrow \text{Denoise}(\mathbf{x}_k(\tau), \hat\epsilon)$
   \STATE $\mathbf{x}_{k-1}\sim\mathcal{N}(\mu_{k-1}, \alpha\Sigma_{k-1})$
   \ENDFOR
   \STATE Extract $r_t$ from $\mathbf{x}_0(\tau)$
\end{algorithmic}
\end{algorithm}
\fi

\begin{algorithm}[h!]
   \caption{The stitching algorithm of DITS (detailed version).
   An abridged version is available in Algorithm~\ref{alg:TSKD_skeleton}.}
   \label{alg:TSKD}
   
   {\bfseries Input:} An inverse dynamics model $f_\phi$, a DITS trajectory generator $D_\theta$, an ensemble of dynamics models $\{\hat p^i_\xi(s_{t+1}|s_t)\}^N_{i=1}$, an acceptance threshold $\tilde{p}$, and a dataset $\mathcal{D}_0$ made up of $T$ trajectories $(\tau_1, ..., \tau_T)$, additional trajectory number $n$, sum of rewards threshold $\lambda$.
   
   \For{$k=0,...,K$}{
   
   Generate $n$ new trajectories $\mathcal{D}'_{k}$ using $D_\theta$ model with Algorithm \ref{alg:DDR-I}
   
   $\mathcal{D}_{k}\leftarrow concat(\mathcal{D}_{k}, \mathcal{D}'_{k})$
   
   Train state-value function $V$ on dataset $\mathcal{D}_k$.
   
   \For{$t=1,...,length(\mathcal{D}_k)$}{
   
   Select $(s, s') = (s_0, s'_0)$ from $\tau_t$
   
   Initialize new trajectory $\hat\tau_t$
   
   \While{not done}{
   
   Create a set of neighbourhood $\{\hat s_j\} = \mathcal{N}(s') \cap \{\text{states in }\mathcal{D}_k\}$
%   \STATE Evaluate dynamics models for the neighbourhood $\hat p_{\xi^i}(\hat s'| s)$
   
   Let $j = \arg \max_i V(\hat s'_i)$
   
   \If{$\min_i\hat p^i_{\xi}(\hat s'_j| s) > \text{mean}_i\hat p^i_{\xi}(s'| s)$
   % , $V(\hat s'_j)=\max_i V(\hat s'_i)$ 
   and $V(\hat s'_j)>V(s')$}{
   
   Generate new action $\tilde a\sim f_\phi(s, \hat s'_j)$
   
   Generate new reward $\tilde r\sim D_{\theta}(s, \hat s'_j)$ via Algorithm \ref{alg:reward_gen}
   
   Add $(s, \tilde a, \tilde r, \hat{s}'_j)$ to new trajectory $\hat \tau_t$
   
   Set $s=\hat s'_j$, and $s'$ to the next state on the trajectory of $\hat s'_j$
   }
   \Else{
   
   Add original transition, $(s, a, r, s')$ to the new trajectory $\hat \tau_t$
   
   Set $s=s'$, and $s'$ to the next state on the trajectory of $s$
   }
   }
   
   \If{$\sum_{i \in \hat \tau_t}r_i \le (1 + \tilde p) \sum_{i \in \tau_t}r_i$}{
   $\hat \tau_t = \tau_t$
   }
   
   }
   
   Sort the trajectories with the values of $\sum_{i \in \hat \tau_t}r_i$
   
   $\mathcal{D}_{k+1} \gets$ the top $T$ trajectories from the sorting result 
   }
   
   $\mathcal{D}_{Aug}\leftarrow$
   Select the trajectories $\tau$ in the dataset $\mathcal{D}_{k+1}$ satisfying $\sum_{i \in \tau} r_i > \lambda$
   
   Train downstream RL models on dataset $\mathcal{D}_{Aug}$   
\end{algorithm}
\clearpage

\subsection{Model-based Trajectory Stitching (MBTS)}
\label{sec:app_algo_details_MBTS}

We next recap, in Algorithm~\ref{alg:MBTS}, the Model-based Trajectory Stitching (MBTS) method from \citep{Hepburn24}.

\begin{algorithm}[H]
   \caption{Model-based Trajectory Stitching (MBTS)}
   \label{alg:MBTS}
   
   {\bfseries Input:} a dataset $\mathcal{D}_0$ with $T$ trajectories $(\tau_1, ..., \tau_T)$
   
   \For{$k=0,...,K$}{
   
   Train the state-value function $V$ on dataset $\mathcal{D}_k$.
   
   \For{{\color{blue} {$t=1,...,T$ (i.e., length of $\mathcal{D}_k$)}}}{
   
   $(s, s') \gets (s_0, s'_0)$ from $\tau_t$. 
   
   Initialize a new trajectory $\hat\tau_t$
   
   \While{not done}{
   % \STATE Create a set of neighbourhood $\{\hat s_j\} = \mathcal{N}(s') \cap \{\text{states in }\mathcal{D}_k\}$
%   \STATE Evaluate dynamics models for the neighbourhood $\hat p_{\xi^i}(\hat s'| s)$
   % \STATE Let $j = \arg \max_i V(\hat s'_i)$
   \If{a state $\hat s'$ in $\mathcal{D}_k$ (possibly in a different trajectory than $t$) is close to $s'$ and is good to stitch to}{
   
   Generate action $\tilde a$ based on $(s, \hat s')$ by IDM
      
   Generate reward $\tilde r$ under $(s, \hat s', \tilde a)$ by {\color{blue} WGAN}
   
   Add $(s, \tilde a, \tilde r, \hat{s}')$ to the new trajectory $\hat \tau_t$
   
   $s \gets \hat s'$
   
   $s' \gets $ the next state on the trajectory of $\hat s'$
   }
   \Else{
   Add original transition to $\hat \tau_t$ and slide $(s, s')$
   }
   }
   % \STATE \textbf{if} $\sum_{i \in \hat \tau_t}r_i \le (1 + \tilde p) \sum_{i \in \tau_t}r_i$}
   % \textbf{then} $\hat \tau_t \gets \tau_t$
   }
   {\color{blue} $\mathcal{D}_{k+1} \gets ( \hat \tau_1, \ldots,  \hat \tau_T)$}
   \hfill $\rhd \ \mathcal{D}_k$ has constant length $T$
   }
   % \STATE {\bfseries Return:} dataset $\mathcal{D}_{K+1}$ for  Behavioural Cloning training
\end{algorithm}

\subsection{Generating reward by training a separate diffuser}
\label{Sec:app_gen_reward}

We also tried to impute the reward by learning another diffusion model,
defining the states as
\begin{align}
\label{eq:xktau_def_ddr_2}
    \mathbf{x}_k(\tau) &\coloneqq (s_t, s_{t+1}, r_t, 
    s_{t+1}, s_{t+2}, r_{t+1}, \ldots, s_{t+H-1}, s_{t+H}, r_{t+H-1})_k \ \in \mathbb{R}^{{\color{blue}H(2m+1)}},
\end{align}
%
% In this case, we will propose a different output block which can be represented as:
% \begin{equation}
%     B^{out}_t\coloneqq(s_t, s_{t+1}, r_t)
% \end{equation}
% while for the initial input, we still replace the reward by a dummy value at the start of the diffusion for a part of the trajectory with the length of $H$:
% \begin{equation}
%     B^{initial}_t\coloneqq(s_t, s_{t+1}, \emptyset)
% \end{equation}
% The structure of the series of output blocks $\mathbf{x}_k(\tau)$ remains unchanged compares to the DDR-I model. 
%
for $k \in [1, K]$.
During the diffusion process, 
we introduce the given $s_t$ and $s_{t+1}$ by clamping the first two entries of $\mathbf{x}_k(\tau)$ for all diffusion steps $k$.
We do not enforce that the forth generated entry must be equal to the second one as in \Eqref{eq:xktau_def_ddr_2},
although they are obviously equal at training phase.
It is also possible to take into account the state of the past $C$ number of steps.
However, in practice we noticed that this does not provide much improvement, 
suggesting that $s_t$ and $s_{t+1}$ are sufficient to predict $r_t$.
Both the training and generation processes are similar to those of the trajectory generator,
and the latter process is detailed in Algorithm~\ref{alg:DDR-II_generation}.

\begin{algorithm}[H]
   \caption{Reward generation by using a separate diffuser with states in \Eqref{eq:xktau_def_ddr_2}}
   \label{alg:DDR-II_generation}
   
   {\bfseries Input:} Noise model $\epsilon_{\theta}$, guidance scale $\omega$, condition $\mathbf{y}$, current state $s$, stitching candidate $s'$
   
   Initialize $\mathbf{x}_K(\tau)\sim\mathcal{N}(\mathbf{0}, \alpha\mathbf{I})$ in $\mathbb{R}^{{\color{blue}H(2m+1)}}$ from \Eqref{eq:xktau_def_ddr_2}
   
   \For{$k = K...1$}{
   
   $\mathbf{x}_k(\tau)[0].\text{states} \gets (s, s')$.
   Set $\hat \epsilon$ with Equation \ref{classifier-free guidance}, using $\theta$, $\mathbf{x}_k(\tau)$, $\mathbf{y}$, and $k$.
   % $\hat{\epsilon}\leftarrow\epsilon_{\theta}(\mathbf{x}_k(\tau), \emptyset, k) + \omega(\epsilon_{\theta}(\mathbf{x}_k(\tau), \mathbf{y}, k) - \epsilon_{\theta}(\mathbf{x}_k(\tau), \emptyset, k))$
   %
     
   Sample $\mathbf{x}_{k-1}(\tau)\sim\mathcal{N}(\mu_{k-1}, \alpha\Sigma_{k-1})$,
   where $(\mu_{k-1}, \Sigma_{k-1})\leftarrow \text{Denoise}(\mathbf{x}_k(\tau), \hat\epsilon)$.
   
   }
   {\bfseries Return:} $\mathbf{x}_0(\tau)[0].\text{reward}$,
   i.e., the $r_t$ in \Eqref{eq:xktau_def_ddr_2}
\end{algorithm}

\subsection{Value Function Criterion}
\label{sec:value_func}

In order to measure if a state is worth stitching, 
a common approach trains a value function over $\mathcal{D}$. 
As the states on the DDR-generated trajectories tend to have a high value, 
they enjoy a higher chance of stitching,
hence promoting these states to be blended into the original dataset.

We approximate the value function using an MLP with parameter $\beta$.
The Bellman error on $\mathcal{D}$ is% \citep{Sutton98}:
\begin{equation}
    \mathcal{L}_V(\beta) := \mathbb{E}_{s, r, s'\sim\mathcal{D}} [(r+\gamma V_\beta(s')-V_\beta(s))^2].
\end{equation}
% Since $V_\beta$ is applied only to in-distribution states, 
% it avoids the out-of-distribution issue when evaluating value function.

% \subsection{Can we share the generators for trajectory and reward?}
% \label{sec:why_two_generators}

% It may appear cumbersome to train two diffusers.
% In particular, since the reward diffuser models $(s_t, s_{t+1}, r_t, s_{t+1}, s_{t+2}, r_{t+1})$ as in \eqref{eq:xktau_def_ddr_2},
% it could also be used to generate the full trajectory which only requires modeling $(s_t, r_t, s_{t+1}, r_{t+1})$ as in \eqref{eq:xktau_def_ddr_1}.
% However, the dimensionality of the former diffusion variables is $H(2m+1)$, 
% \textbf{doubling} the latter which is $H(m+1)$.
% %
% Given a fixed amount of data,
% the (larger) reward diffuser is generally less accurately trained than 
% the (smaller) trajectory diffuser.
% For example, in hopper medium expert,
% the former generated trajectories that yielded normalized return of $93.7 \pm 5.8$,
% while the latter improved it to $111.5 \pm 0.5$.
% Therefore, we would prefer to restrict the use of the reward diffuser only to necessary.

% Since we need to generate a reward $r$ for a candidate transition from the current state $s$ to a next state $s'$, DDR-I is insufficient since it only utilizes the knowledge of current state $s$. 

\section{Hyperparameter and Architectural Details}
\label{hyperparams}
In this section, we describe the hyperparameter choice for our model and the architectural details for our model:

\begin{itemize}
\item We use a temporal U-Net \citep{Janner22} for the noise $\epsilon_\theta$ modeling. It consists a U-Net structure with 6 repeated residual blocks, while each block consisting two temporal convolutions, each followed by group norm \citep{Wu18}, and a final Mish nonlinearity \citep{Misra19}. Both of the timestep and condition embeddings are 128-dimensional vectors produced by two separate 2-layered MLP with 256 hidden units and Mish nonlinearity. The embeddings are concatenated together before getting added to the activations of the first temporal convolution within each block. Our code for the DDR model is a modification of the code for the original Decision Diffuser \citep{Ajay23}, for which the link is \url{https://github.com/anuragajay/decision-diffuser}.
\item We use a 2-layered MLP with 512 hidden units and ReLU activations for the modeling of the inverse dynamics model $f_\phi$.
\item We train $\epsilon_\theta$ and $f_\phi$ using the Adam optimizer \citep{Kingma15} with a learning rate $2 \cdot 10^{-4}$ and batch size of 32 for $1e6$ training steps.
\item We choose the probability $p$ of removing the conditioning information to be 0.25.
\item We use $K=200$ diffusion steps.
\item We choose context length $C\in\{1, 20\}$, $C=20$ is preferred in the Kitchen datasets. Both $C$ values are able to generate decent results in the locomotion datasets, but $C=20$ tends to have more stability.
\item We use a planning horizon $H$ with $H-C=100$ in all the D4RL locomotion tasks, while using $H-C=56$ in D4RL kitchen tasks.
\item We use a guidance scale $\omega\in\{1.2,1.4,1.6,1.8\}$ but the exact choice varies by task.
\item We choose $\alpha=0.5$ for low temperature sampling.
\item We use a 3-layered MLP with 300 hidden units with ReLU activation to model the foward dynamics. The network takes a state $s$ for input and output a mean $\mu$ and a standard deviation $\sigma$ for a Gaussian distribution $\mathcal{N}(\mu,\sigma^2)$. For all experiments, an ensemble size of 7 is used with the best 3 being chosen. We train the forward dynamics models with Adam optimizer \citep{Kingma15}, a learning rate of $3 \cdot 10^{-4}$ and a batch size of 256. We initialize the parameter with different random seeds for each forward dynamics model in the ensemble.
\item We choose to use a 2-layered MLP with 256 hidden units and a ReLU activation to parameterize the value function. We train two value functions with different parameter initialization and take the minimum of the two during the stitching process. We train the model with Adam optimizer and a learning rate of $3 \cdot 10^{-4}$ and a batch size of 256.
\item We choose the neighbourhood radius $\rho\in\{0.1,1.0,3.0\}$ while the exact choice varies by task.
\item We use the additional trajectory number $n=300$ for each epoch during stitching.
\item We use the sum of rewards threshold $\lambda = \max_t(\sum_{\tau_t}r^i) - \kappa$, while $\kappa$ is in the range of $[500,1000]$ depending on the task.
\item We choose the acceptance threshold $\tilde p=0.1$ to ensure the stitched trajectory only to be used when a significant improvement is guaranteed.
\item We use a behavioural cloning model parameterized by a 2-layered MLP with 256 hidden units and ReLU activation.
\item We choose to use the epoch number $K=1, 2, 3, 4$. For most cases, using $K=2$ will have the sufficient performance gain, and the results will saturate in the following epochs.
\end{itemize}

\subsection{Neighborhood Selection}
\label{sec:neighborhood}

For trajectory stitching, we discourage stitching two states that are far away. 
In this case, stitching in a created neighbourhood for the next state is preferred. 
Multiple metrics are available to define the neighbourhood by measuring the distance in the state space. 
\citet{Castro20} proposed a pseudo-metric,
but its computational cost is rather high.
Therefore, we resort to a more straightforward approach by applying the L2 norm for neighbourhood selection:
\begin{equation}
    \mathcal{N}(s)\coloneqq\{\hat s:  \Vert \hat s - s \Vert_2 < \rho \},
\end{equation}
where $\rho$ is a hyperparameter with a relatively small value.

\section{Classifier-free Guidance Details}
In this section, we show the derivation of the classifier-free guidance Eq.\ref{classifier-free guidance} for completeness. From the derivation outlined in prior works \citep{luo2022understanding}, we know that $\nabla_{\mathbf{x}_k(\tau)}\log q(\mathbf{x}_k(\tau)\vert \mathbf{y}(\tau))\propto -\epsilon_\theta(\mathbf{x}_k(\tau), \mathbf{y}(\tau), k)$. Furthermore, we can derive:
$$
q(\mathbf{x}_k(\tau)\vert \mathbf{y}(\tau)) = q(\mathbf{x}_k(\tau))\frac{q(\mathbf{x}_k(\tau)\vert \mathbf{y}(\tau))}{q(\mathbf{x}_k(\tau))}
$$
$$
\Rightarrow \log q(\mathbf{x}_k(\tau)\vert \mathbf{y}(\tau)) = \log q(\mathbf{x}_k(\tau)) + (\log q(\mathbf{x}_k(\tau)\vert \mathbf{y}(\tau)) - \log q(\mathbf{x}_k(\tau)))
$$
In order to sample from $q(\mathbf{x}_0(\tau)\vert \mathbf{y}(\tau))$ with classifier-free guidance, we multiply the second term with conditional guidance factor $\omega$:
$$
\log \hat q \coloneqq \log q(\mathbf{x}_k(\tau)) + \omega (\log q(\mathbf{x}_k(\tau)\vert \mathbf{y}(\tau)) - \log q(\mathbf{x}_k(\tau)))
$$
$$
\Rightarrow\nabla_{\mathbf{x}_k(\tau)}\log \hat q = \nabla_{\mathbf{x}_k(\tau)}\log q(\mathbf{x}_k(\tau)) + \omega (\nabla_{\mathbf{x}_k(\tau)}\log q(\mathbf{x}_k(\tau)\vert \mathbf{y}(\tau)) - \nabla_{\mathbf{x}_k(\tau)}\log q(\mathbf{x}_k(\tau)))
$$
$$
\Rightarrow \hat\epsilon \coloneqq \epsilon_\theta(\mathbf{x}_k(\tau), \emptyset, k) + \omega(\epsilon_\theta(\mathbf{x}_k(\tau), \mathbf{y}(\tau), k) - \epsilon_\theta(\mathbf{x}_k(\tau), \emptyset, k))
$$
In this case, we have our Eq.\ref{classifier-free guidance}. This result can also be expanded to a composing form with a number of different conditioning variables \citep{Ajay23}.

\section{Additional Ablations of DITS Components}
\label{sec:ablation_DITS}
\subsection{Ablation on Generation and Stitching}

\begin{figure*}[t]
\begin{minipage}[t]{0.32\textwidth}
\begin{figure}[H]
\centering
\includegraphics[scale=0.335]{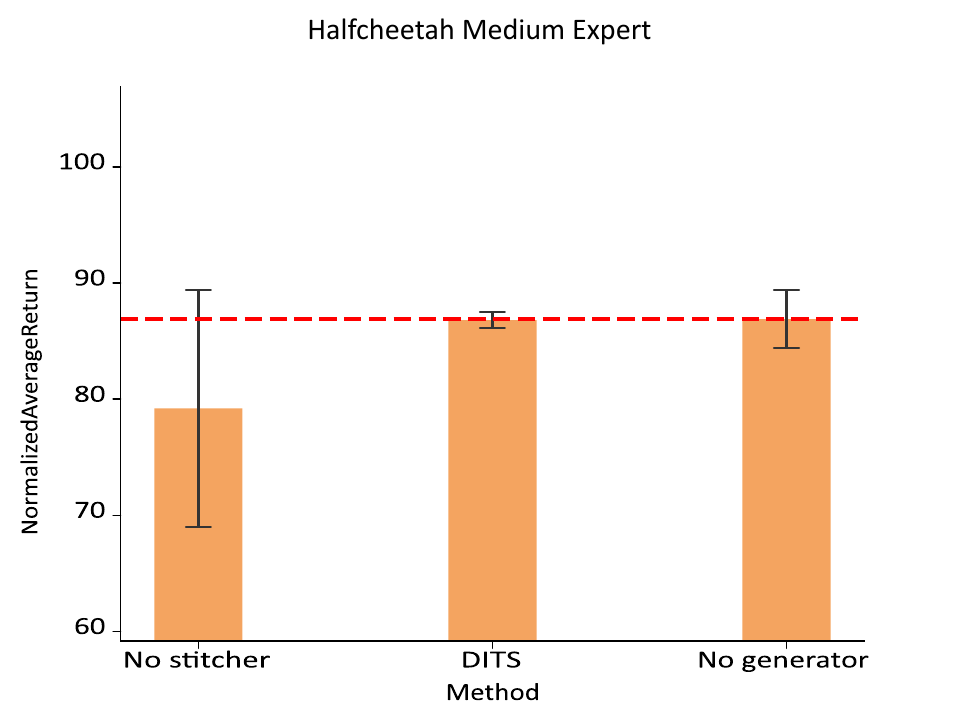}
\end{figure}
\end{minipage}
\hfill
\begin{minipage}[t]{0.32\textwidth}
\begin{figure}[H]
\centering
\includegraphics[scale=0.335]{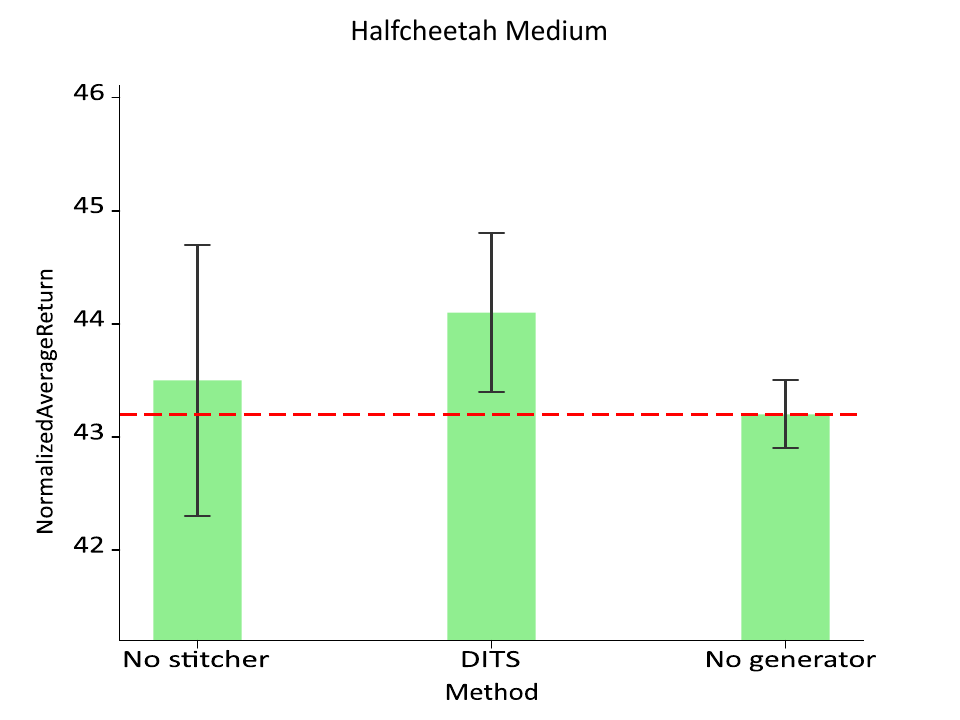}

\end{figure}
\end{minipage}
\hfill
\begin{minipage}[t]{0.32\textwidth}
\begin{figure}[H]
\centering
\includegraphics[scale=0.335]{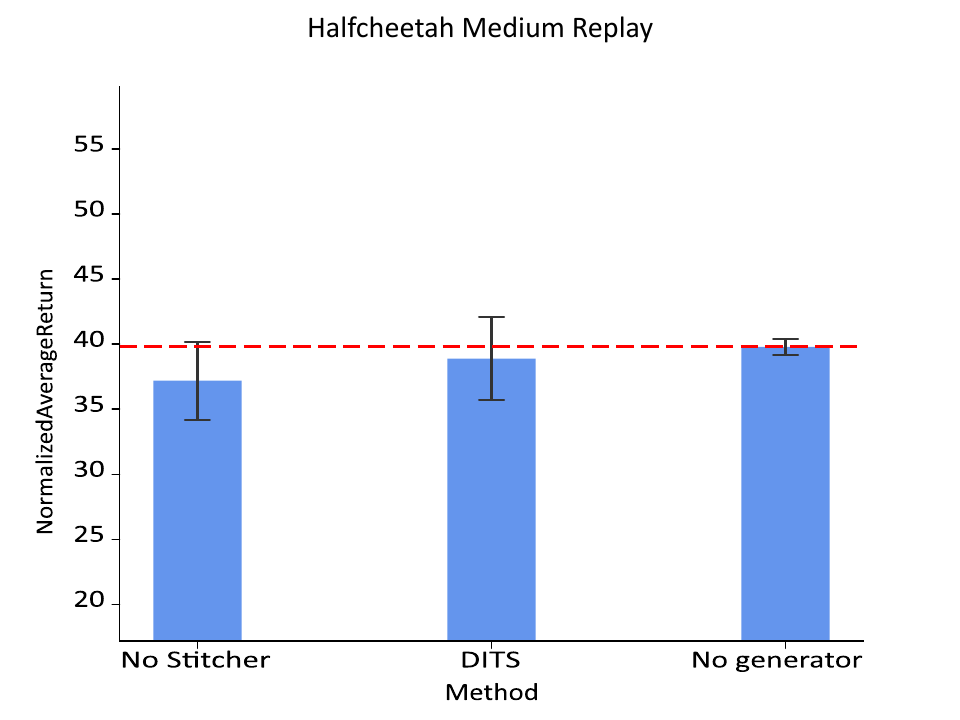}
\end{figure}
\end{minipage}
\begin{minipage}[t]{0.32\textwidth}
\begin{figure}[H]
\centering
\includegraphics[scale=0.335]{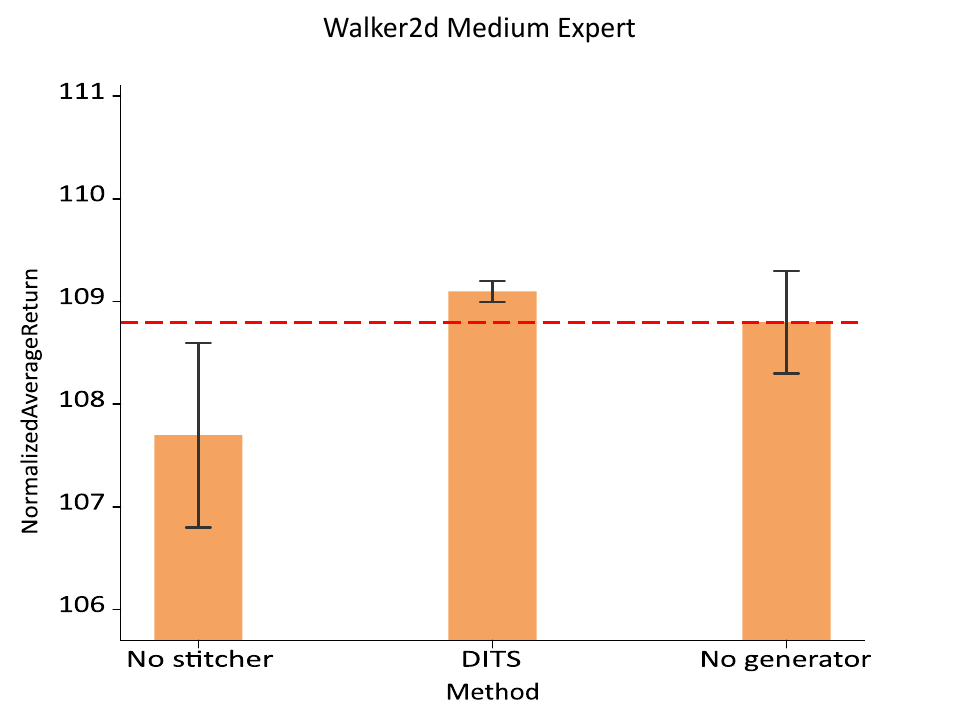}
\end{figure}
\end{minipage}
\hfill
\begin{minipage}[t]{0.32\textwidth}
\begin{figure}[H]
\centering
\includegraphics[scale=0.335]{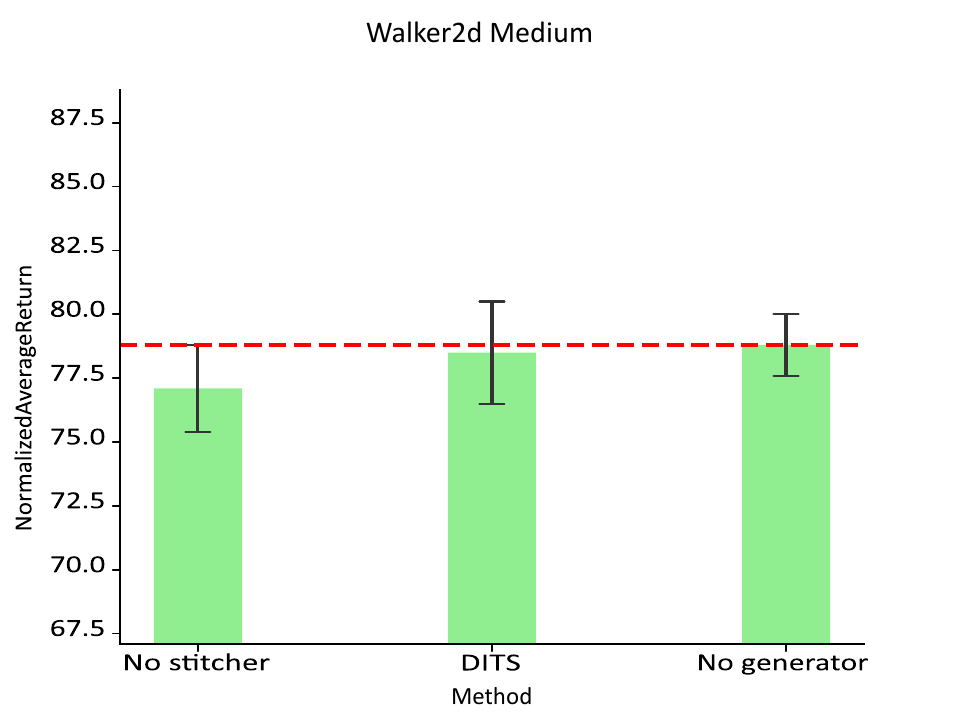}

\end{figure}
\end{minipage}
\hfill
\begin{minipage}[t]{0.32\textwidth}
\begin{figure}[H]
\centering
\includegraphics[scale=0.335]{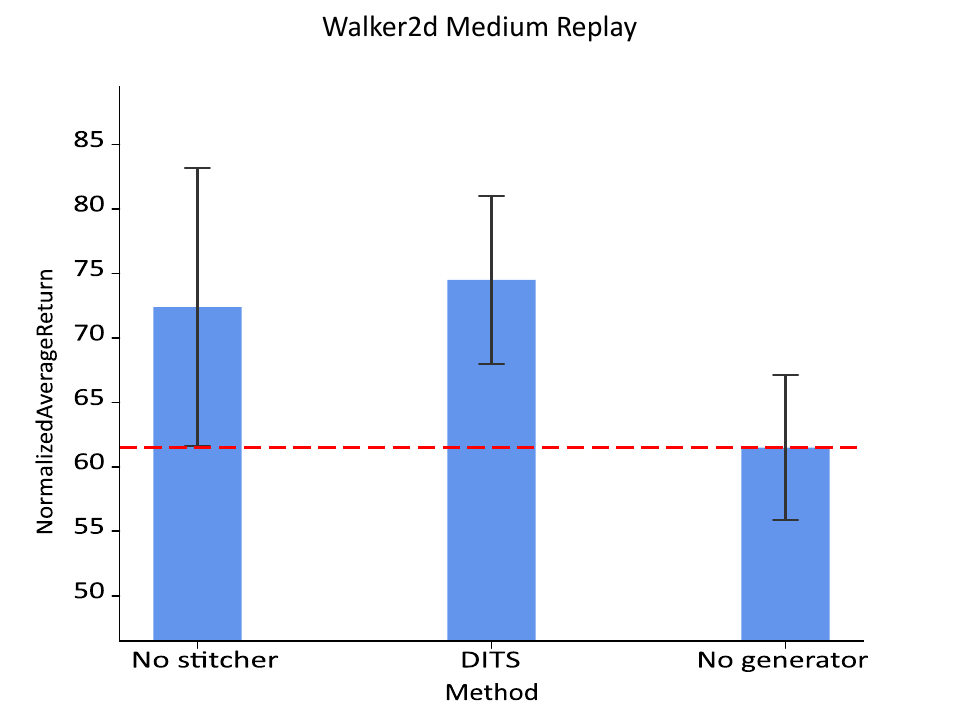}
\end{figure}
\end{minipage}
\vspace{-0.4em}

\caption{\textbf{Ablation study on trajectory generation and stitching}.
The setting is the same as Figure~\ref{ablation-hopper},
except that the datasets are Halfcheetah (top) and Walker2d (bottom).
Difficulty levels are medium expert, medium, and medium replay (left to right).
}
\label{ablation-addition}
\vspace{-1em}
\end{figure*}

We performed ablation studies on the Halfcheetah and Walker2d environments,
in addition to the ablation study in Section~\ref{ablation_TSKD}. 
The settings are the same as there,
and the results are shown in Figure~\ref{ablation-addition}. 
The observations are similar to those in Section~\ref{ablation_TSKD}.

% We do extra experiments for trajectory generation and stitching. We can still observe improvements by introducing the generator part and the stitcher part in most of the situations. We evaluate the ablation on BC policy. The abaltion results are shown in Figure \ref{ablation-addition}. In most of the cases, our method matches and outperforms the original MBTS method.

% \subsection{Ablation on Conditional Guidance Scale}
% We analyzed the sensitivity of the conditional guidance scale $\omega$ of the conditional diffusion model on Hopper Medium Expert dataset. We evaluate the generation quality via computing the correlation between expert data. We demonstrate the results in Table \ref{tab:conditional-guidance-ablation}.

% \begin{table*}[ht]
% \setlength\tabcolsep{3pt}
% % \vskip 0.15in
% \begin{center}
% \begin{small}
% \begin{tabular}{lcccr}
% \toprule
% Conditional Guidance Scale  & 1.2 & 1.4 & 1.6 & 1.8\\
% \midrule
% Correlation& 0.996  & 0.996 & 0.993 & 0.989\\
% \bottomrule
% \end{tabular}
% \end{small}
% \end{center}
% \vskip -0.1in
% \caption{\textbf{Ablation on Conditional Guidance} We ablate over the conditional guidance scale in Hopper Medium Expert. We observe better generation quality when the scale is smaller for datasets with relatively high quality (contains a portion of expert data).}
% \label{tab:conditional-guidance-ablation}
% \end{table*}

\section{Assessment of Reward Prediction}
\label{sec:reward_assess}
\begin{figure}[H]
\vskip 0.1in
\begin{center}
\centerline{\includegraphics[scale=0.35]{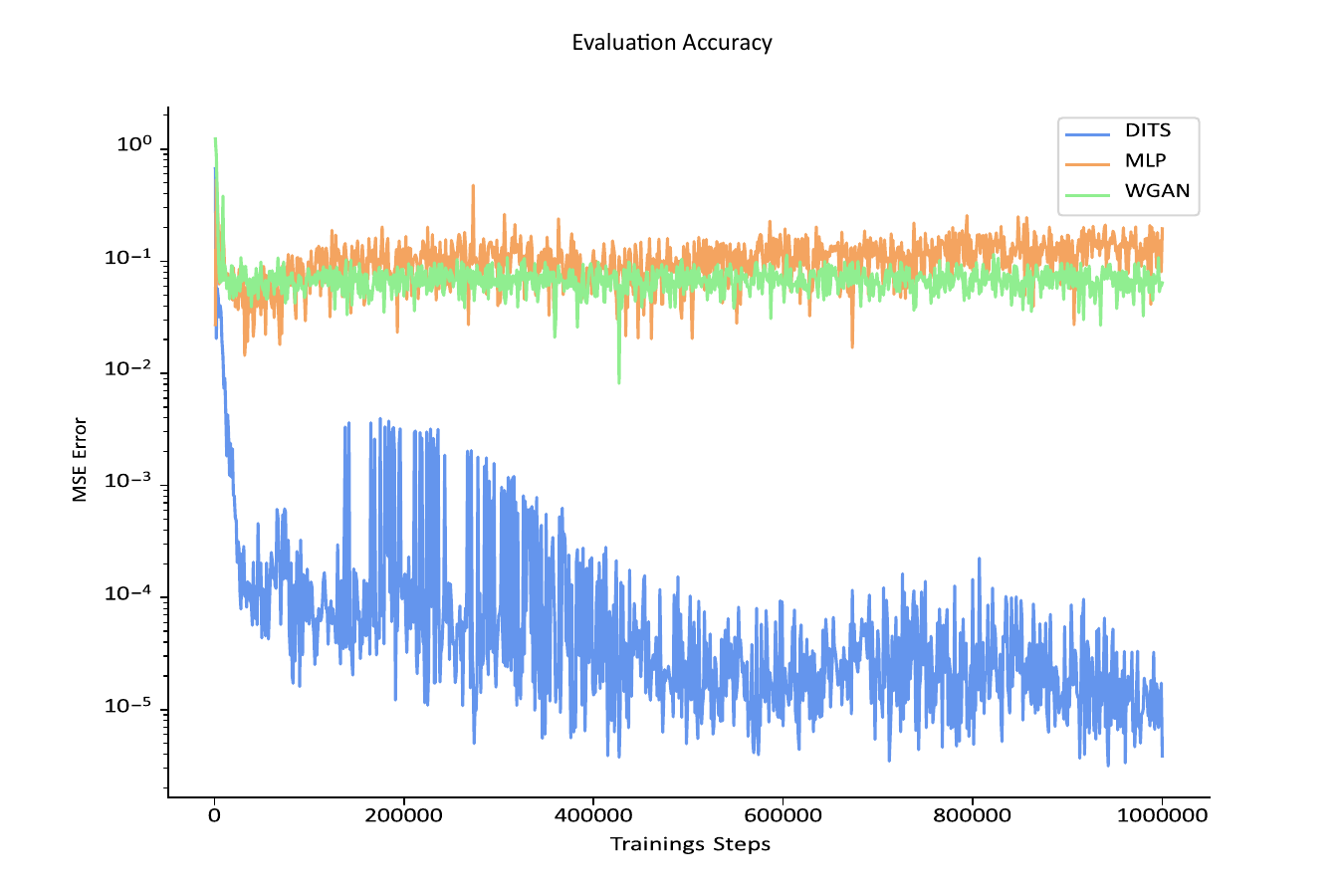}}
\caption{\textbf{The reward generation accuracy analysis}. We trained the DITS trajectory generation model on the Hopper Medium Expert dataset and evaluated the MSE error of the DITS model's generation on the environment during the training process. We compared our results with an MLP generator and a WGAN generator.}
\label{MSE}
\end{center}
\vskip -0.1in
\end{figure}

In this section, we analyze the reward generation ability of the DITS model and compare our model with an MLP generator and WGAN generator to show that our model has a better reward generating performance. We evaluate the mean MSE error of the predicted reward and the true reward over the evaluating batch $\mathcal{D}_B$ via the metric:
\begin{equation}
    M = \mathbb{E}_{s,s',r\sim \mathcal{D}_B}(D_{\theta}(s, s')-r)^2
\end{equation}
We compare our generator with 2-layered MLP generator with 256 hidden units and a WGAN with 2-dimensional latent space and 256 hidden units on the D4RL Hopper Medium-Expert dataset. The results are shown in Figure \ref{MSE}.

\end{document}